\pdfoutput=1
\documentclass[11pt]{article}

\usepackage[preprint]{acl}

\usepackage{amsmath}
\usepackage{amsfonts}
\usepackage{times}
\usepackage{latexsym}
\usepackage{amssymb}
\usepackage{algorithm}
\usepackage{algpseudocode}
\usepackage{makecell}
\usepackage[T1]{fontenc}
\usepackage[table]{xcolor}

\definecolor{lightgray}{gray}{0.4}
\definecolor{darkblue}{RGB}{94,110,186}
\definecolor{darkGreen}{RGB}{92, 148, 110}
\definecolor{darkgreen}{RGB}{0,120,0}
\definecolor{lightorange}{RGB}{255,160,0}
\definecolor{darkred}{RGB}{139,0,0}
\definecolor{yellow}{RGB}{255,255,230}
\definecolor{cashew}{RGB}{255,216,158}

\usepackage[utf8]{inputenc}

\usepackage{microtype}

\usepackage{inconsolata}

\usepackage{graphicx}
\usepackage{booktabs}
\usepackage{graphicx}
\usepackage{multirow}
\usepackage{url}
\usepackage{array}
\usepackage{stfloats}

\usepackage{listings}
\usepackage{tcolorbox}
\tcbuselibrary{listings, skins, breakable}
\lstdefinelanguage{json}{
    basicstyle=\ttfamily\footnotesize,
    numbers=none,
    numberstyle=\tiny,
    stepnumber=1,
    numbersep=5pt,
    showstringspaces=false,
    breaklines=true,
    frame=single,
    backgroundcolor=\color{gray!5},
    string=[s]{"}{"},
    morestring=[b]',
    literate=
     *{0}{{{\color{blue}0}}}{1}
      {1}{{{\color{blue}1}}}{1}
      {2}{{{\color{blue}2}}}{1}
      {3}{{{\color{blue}3}}}{1}
      {4}{{{\color{blue}4}}}{1}
      {5}{{{\color{blue}5}}}{1}
      {6}{{{\color{blue}6}}}{1}
      {7}{{{\color{blue}7}}}{1}
      {8}{{{\color{blue}8}}}{1}
      {9}{{{\color{blue}9}}}{1}
}

\usepackage{xcolor}
\usepackage[normalem]{ulem}

\definecolor{AlertRed}{HTML}{ED1C24}
\definecolor{SuccessGreen}{HTML}{248B24}
\definecolor{AzureBlue}{HTML}{0685DD}

\title{~\raisebox{-1.3ex}{\includegraphics[height=1.8em]{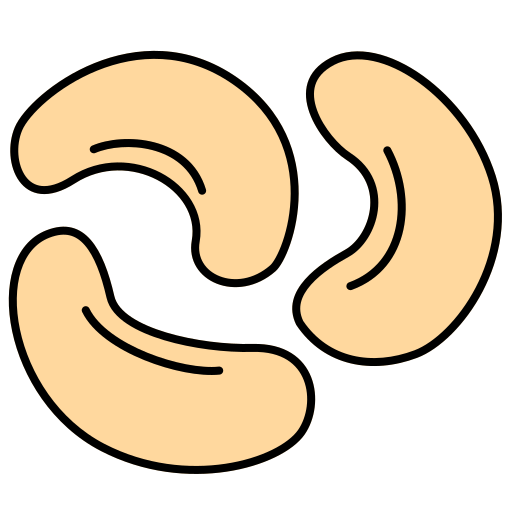}}\hspace{0.3em}CASHEW: Stabilizing Multimodal Reasoning via\\ Iterative Trajectory Aggregation}

\author{Chaoyu Li$^1$, Fei Tao$^2$, Pooyan Fazli$^1$ \\
    $^1$Arizona State University, $^2$NewsBreak\\
         \texttt{\{chaoyuli, pooyan\}@asu.edu}, 
         \texttt{fei.tao@newsbreak.com}}

\begin{document}

\maketitle

\begin{abstract}
  Vision-language models achieve strong performance across a wide range of multimodal understanding and reasoning tasks, yet their multi-step reasoning remains unstable.\ Repeated sampling over the same input often produces divergent reasoning trajectories and inconsistent final predictions.\ To address this, we introduce two complementary approaches inspired by test-time scaling: (1) \textbf{\textsc{Cashew}}, an inference-time framework that stabilizes reasoning by iteratively aggregating multiple candidate trajectories into higher-quality reasoning traces, with visual verification of object presence providing grounding cues during aggregation, and (2) \textbf{\textsc{Cashew-RL}}, a learned variant that internalizes this aggregation behavior within a single model.\ \textsc{Cashew-RL} is trained using Group Sequence Policy Optimization (GSPO) with a composite reward that encourages correct answers grounded in minimal yet sufficient visual evidence, while adaptively allocating reasoning effort based on task difficulty.\ This training objective enables robust self-aggregation at inference. Extensive experiments on 13 image understanding, video understanding, and video reasoning benchmarks show significant performance improvements, with \textsc{Cashew}/\textsc{Cashew-RL} achieving average gains of +4.1/+5.3 percentage points on Qwen3-VL-4B and +2.6/+3.0 percentage points on Qwen3-VL-8B, respectively.
\end{abstract}

\section{Introduction}
\label{sec:intro}

Vision-language models (VLMs) have become a dominant paradigm for multimodal understanding, enabling unified models that reason jointly over images, videos, and text~\cite{zhang2024llavavideo, bai2025qwen25vl, zhang2025videollama3}. While recent VLMs achieve strong performance on tasks such as visual question answering and multimodal dialogue, their reasoning processes remain fragile. Multi-step predictions are often sensitive to sampling noise, prone to visual hallucinations~\cite{Li_2025_CVPR}, and inconsistent across inference runs~\cite{li2023pope}. This instability is especially evident in complex reasoning scenarios, where small perceptual or interpretive errors can propagate through longer chains of thought.

\begin{figure}[t]
    \centering
    \includegraphics[width=1\linewidth]{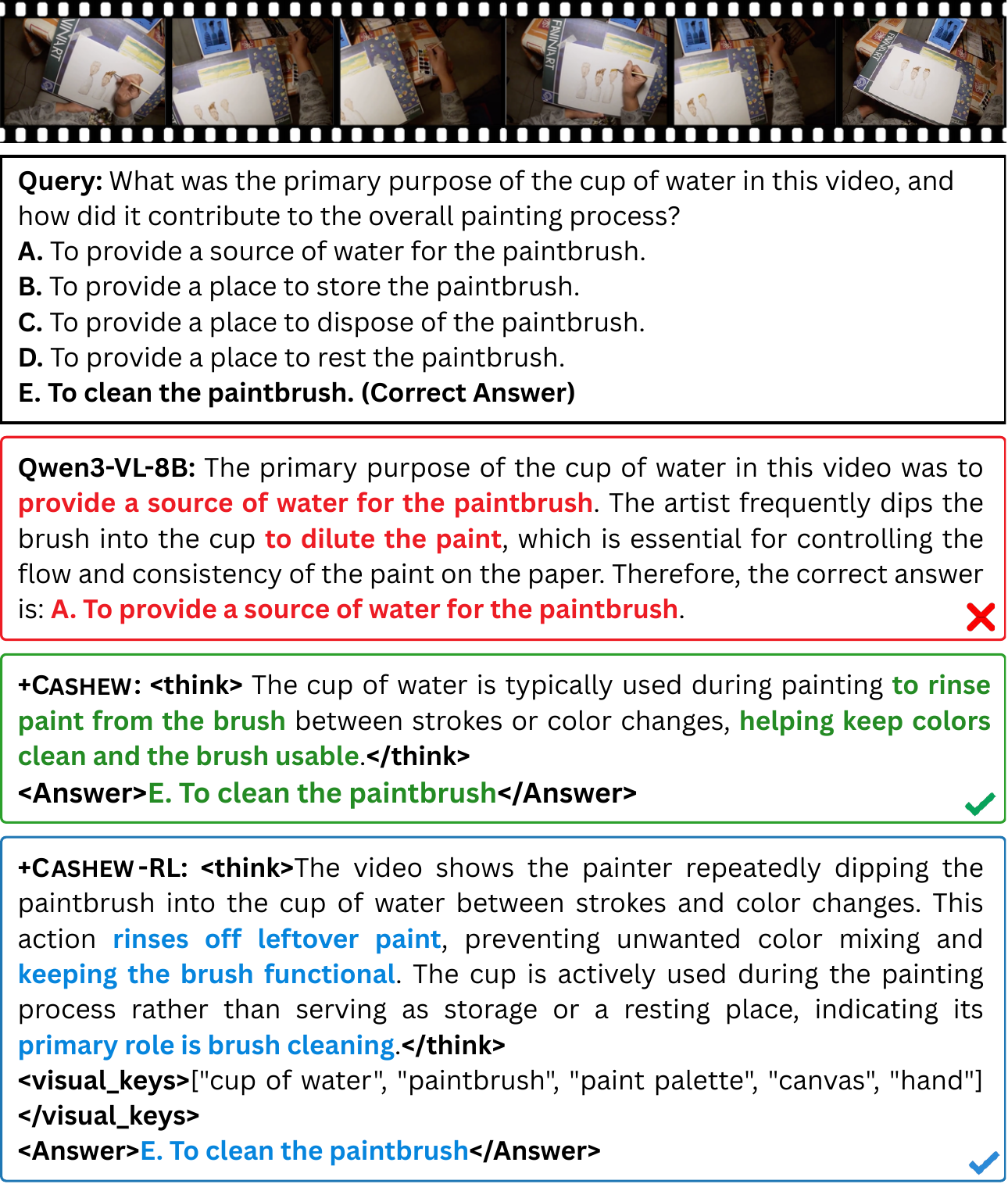}
    \vspace{-7mm}
    \caption{\footnotesize\textbf{\textsc{Cashew} enables robust reasoning through visually grounded iterative aggregation.} Unlike standard vision-language models that rely on single-path reasoning and are prone to hallucinations, \textsc{Cashew} aggregates multiple reasoning trajectories with explicit visual verification. \textbf{\textsc{Cashew-RL}} further internalizes this aggregation behavior via reinforcement learning.}
    \label{fig:teaser}
    \vspace{-5mm}
\end{figure}

In large language models, test-time scaling has emerged as an effective strategy to mitigate such instability by allocating additional inference-time computation, such as sampling multiple reasoning trajectories or extending deliberation depth, to obtain more reliable outputs~\cite{wang2023selfconsistency, wang2025selfconsistency, chen2023universalselfconsistency}. Inspired by this paradigm, recent multimodal methods adopt iterative inference mechanisms that repeatedly refine spatial-temporal attention and textual predictions across multiple passes to improve reasoning consistency~\cite{yan2025videochatr15}. However, most existing multimodal test-time scaling approaches follow a ``sample-and-select'' paradigm. They generate multiple independent reasoning chains and attempt to choose the best one, discarding the partial insights contained in rejected trajectories. Moreover, simply increasing reasoning length or sampling count in VLMs carries a fundamental risk. Without explicit grounding, early perceptual errors can be amplified rather than corrected, leading models to reason more without reasoning better. In other words, these methods encourage models to think \textit{longer}, but not necessarily to think \textit{together} or to verify their conclusions against visual evidence.

To address these limitations, we propose \textsc{Cashew} (\textbf{C}andidate \textbf{A}ggregation and \textbf{S}ynt\textbf{H}esis of \textbf{E}vidence and re\textbf{W}ards), a framework for visually grounded iterative aggregation. Instead of selecting a single trajectory, \textsc{Cashew} treats reasoning as an evolutionary process: at each iteration, it synthesizes a population of candidate trajectories into a higher-quality aggregate. To prevent hallucinations, object-level claims are verified against the visual input, and grounded evidence guides subsequent aggregation steps. This ensures the resulting consensus is anchored in visual evidence.\ Building on this, we introduce \textsc{Cashew-RL}, a learned variant that internalizes aggregation during post-training. \textsc{Cashew-RL} allows the VLM to integrate multiple reasoning trajectories with visual evidence through learned parameters, thereby improving aggregation efficiency under test-time sampling. The model is trained using Group Sequence Policy Optimization (GSPO) with a composite reward that encourages correct answers grounded in sufficient visual evidence while adaptively allocating reasoning effort based on task difficulty. In summary, our contributions are:

\begin{itemize}
  \item[$\bullet$] \textsc{Cashew}: an inference-time framework that stabilizes multimodal reasoning by iteratively aggregating candidate trajectories with visual verification of object presence providing auxiliary grounding cues.
  \item[$\bullet$] \textsc{Cashew-RL}: a learned variant that internalizes trajectory aggregation during post-training via GSPO and a composite reward, enabling visually grounded aggregation of multiple reasoning trajectories with adaptive control of reasoning effort.
  \item[$\bullet$] Extensive experiments on 13 benchmarks and multiple backbone families demonstrate consistent and significant performance improvements, with Cashew/Cashew-RL achieving average gains of +4.1/+5.3 percentage points on Qwen3-VL-4B and +2.6/+3.0 percentage points on Qwen3-VL-8B, respectively.
\end{itemize}

\section{Related Work}
\label{sec:related-work}
 
\paragraph{Multimodal Reasoning.} Vision-language models (VLMs)~\cite{alayrac2022flamingo, li2023blip2, liu2024llava, bai2023qwenvl} provide unified architectures for reasoning over images and videos, supporting tasks such as captioning, question answering, and visually grounded dialogue. Recent work further improves multimodal reasoning via explicit chain-of-thought prompting and process-level supervision~\cite{chen2024visualcot,zhang2024mmcot}, showing that modeling intermediate reasoning steps outperforms direct answer prediction~\cite{zhou2025proreason}. However, most existing VLMs generate reasoning trajectories independently, without mechanisms for iterative refinement or aggregation across multiple paths. In this work, we propose \textsc{Cashew} and \textsc{Cashew-RL}, two methods that address this limitation by iteratively aggregating multiple reasoning trajectories to produce more reliable and grounded answers.


\paragraph{Test-Time Scaling.}\ 
Test-time scaling improves reasoning without additional training~\cite{wang2023selfconsistency, wang2025selfconsistency} by allocating more inference-time computation to produce consistent outputs~\cite{shinn2023reflexion, madaan2023selfrefine, yao2023treethoughts, wu2023vguided, luo2025coherentmultimodalreasoningiterative, yuan2025cimrcontextualizediterativemultimodal}. LongPerceptualThoughts~\cite{liao2025longperceptualthoughts} extends reasoning budgets to generate longer, self-corrective chains, while VideoChat-R1.5~\cite{yan2025videochatr15} iteratively refines spatial-temporal attention and textual predictions for video understanding. These approaches increase reasoning depth but still operate on a single trajectory. In contrast, \textsc{Cashew} iteratively aggregates multiple reasoning trajectories into a unified process.

\paragraph{Reinforcement Learning for VLMs.} 
Reinforcement learning aligns multimodal models with human or process-level rewards. Methods such as RLHF~\cite{ouyang2022rlhf}, RLAIF~\cite{lee2024rlaif}, DPO~\cite{rafailov2024dpo}, GRPO~\cite{shao2024grpo}, and GSPO~\cite{zheng2025gspo} improve factuality and reasoning in LLM and VLMs. Recent work~\cite{ong2025vlprm} introduces step-wise rewards to guide intermediate reasoning, while GFlowVLM~\cite{kang2025gflowvlm} models distributions over reasoning trajectories using generative flow networks. In contrast, our RL formulation trains an aggregation policy that fuses and refines multiple reasoning trajectories using correctness and consistency rewards, enabling the model to internalize iterative aggregation.

\begin{figure*}[t]
 \centering
 \includegraphics[width=0.90\linewidth]{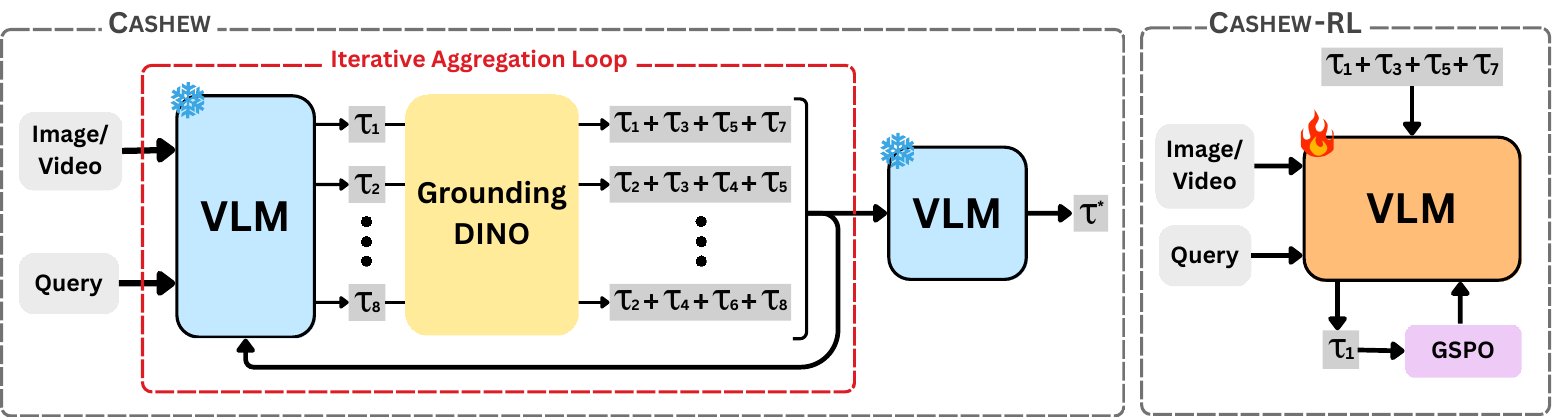}
 \caption{\footnotesize\textbf{Overview of the \textsc{Cashew} and \textsc{Cashew-RL} frameworks.}
\textbf{Left:} \textsc{Cashew} performs test-time iterative aggregation by generating a population of candidate trajectories from a frozen VLM and synthesizing subsets into refined trajectories over multiple iterations to produce a consolidated trajectory $\tau^{*}$. A lightweight visual verification module (instantiated with Grounding DINO in our implementation) provides consistency signals to suppress unsupported object-level claims during aggregation.
\textbf{Right:} \textsc{Cashew-RL} extends this framework via post-training with GSPO, teaching the VLM to internally aggregate multiple candidate trajectories into high-quality, visually grounded reasoning traces.}
\label{fig:method}
\vspace{-2mm}
\end{figure*}

\section{Problem Formulation and Preliminaries}
\label{sec:problem}

Given a multimodal input $\mathbf{x} = {I, q}$, where $I$ is an image or video and $q$ is a textual query, a vision-language model aims to produce both a final answer $y$ and a reasoning trajectory $\tau$. A reasoning trajectory is defined as a sequence of intermediate textual reasoning steps that interpret visual observations and progressively support the predicted answer. Under standard inference, the model generates a single trajectory per input, which can be sensitive to sampling noise and brittle in complex multimodal reasoning scenarios.

Instead, we consider a multi-trajectory inference setting, in which the model produces a set of candidate reasoning trajectories, $\mathcal{T} = \{\tau_1, \ldots, \tau_N\}$, for the same input. Rather than selecting one trajectory in isolation, we formulate multimodal reasoning as an \textit{iterative aggregation problem}, where an aggregation operator progressively maps $\mathcal{T}$ to a refined trajectory $\tau^{*}$. This process integrates complementary information across trajectories, resolves inconsistencies, and emphasizes reasoning steps that are consistent with the visual evidence. The goal is to obtain a reasoning trace that is more stable and visually grounded than any individual trajectory.

Building on this formulation, we present \textbf{\textsc{Cashew}}, an iterative aggregation framework for improving multimodal reasoning in VLMs. \textsc{Cashew} functions as a plug-and-play test-time scaling method that aggregates multiple sampled reasoning trajectories at inference time. We further introduce \textsc{Cashew-RL}, a learned variant of \textsc{Cashew} that internalizes aggregation behavior during post-training using Group Sequence Policy Optimization (GSPO). While \textsc{Cashew} performs aggregation solely at inference time, \textsc{Cashew-RL} learns a robust aggregation policy that enables the model to combine reasoning trajectories and verified visual evidence through learned parameters, improving aggregation efficiency under test-time sampling. Further details on data generation are provided in Appendix~\ref{app:data}.

\section{\textsc{Cashew}~\raisebox{-1.0ex}{\includegraphics[height=1.6em]{img/cashew.png}}}
\label{sec:Cashew}

\textsc{Cashew} is a test-time framework that models reasoning as an evolving population of candidate trajectories that are iteratively synthesized and refined. We describe its stages below, with pseudocode provided in Appendix~\ref{app:algo}.

\paragraph{Population Initialization.} Let $\mathcal{T}_t$ denote the population of candidate reasoning trajectories at iteration $t$. The base VLM $p_\theta$ first produces an initial population of $N$ candidate trajectories:
\begin{equation}
  \mathcal{T}_1 = \{\tau_i^{(1)} \sim p_\theta(\cdot \mid \mathbf{x})\}_{i=1}^{N}.
\end{equation}
Each trajectory $\tau_i^{(1)}$ consists of a reasoning text $r_i^{(1)}$ and a predicted answer $a_i^{(1)}$.

\paragraph{Subset Sampling.} At iteration $t$, to generate the $i$-th trajectory $\tau_i^{(t+1)}$, we sample a random subset of trajectories $S_i^{(t)} \subset \mathcal{T}_t$ of size $M$ as candidates. This sampling strategy introduces diversity and mitigates collapse to a single reasoning mode.

\paragraph{Visual Verification Module.} 
To mitigate hallucination, we incorporate a lightweight visual-text grounding verification step before aggregation. For each candidate trajectory $\tau_j^{(t)} \in S_i^{(t)}$, we extract mentioned objects using a text parser $\mathcal{E}$:
\begin{equation}
    \mathcal{O}_j^{(t)} = \mathcal{E}(r_j^{(t)}).
\end{equation}
Each object $o_k \in \mathcal{O}_j^{(t)}$ is then evaluated by a visual 
verification function $\mathcal{G}$ to assess its presence in the visual input $I$:
\begin{equation}
    v(o_k) = \mathbf{1}\big[\mathcal{G}(I, o_k) > \delta_g\big],
\end{equation}
where $\mathcal{G}(I, o_k)$ produces a visual grounding confidence score and $\delta_g$ is a confidence threshold. In this work, we instantiate 
$\mathcal{G}$ using Grounding DINO~\cite{liu2024groundingdino}, though the framework is agnostic to the specific object verifier and can incorporate alternative object-grounding models. We collect all verified objects as grounded visual evidence:
\begin{equation}
    \mathcal{V}_j^{(t)} = \{o_k \in \mathcal{O}_j^{(t)} \mid v(o_k) = 1\}.
\end{equation}
The verified evidence serves as auxiliary consistency guidance during 
trajectory aggregation rather than as a direct reasoning component.

\paragraph{Grounded Aggregation.}
Finally, the model synthesizes the sampled trajectories and their corresponding visual verification results to generate an updated reasoning trajectory. We explicitly condition the generation on both the reasoning texts and the corresponding verified visual evidence:
\begin{equation}
    \tau_i^{(t+1)} \sim p_\theta\big(\cdot \mid \mathbf{x}, S_i^{(t)}, \{\mathcal{V}_j^{(t)}\}_{\tau_j^{(t)} \in S_i^{(t)}}\big).
\end{equation}
By providing $\{\mathcal{V}_j^{(t)}\}$, the model can distinguish which parts of the candidate reasoning are visually supported, encouraging it to anchor the new trajectory $\tau_i^{(t+1)}$ in verified evidence. The population is then updated as $\mathcal{T}_{t+1} = \{\tau_i^{(t+1)}\}_{i=1}^{N}$.

\paragraph{Final Aggregation.} We perform up to $T-1$ updates, where $T$ includes the initial population, and stop early upon unanimous predictions.
If consensus is reached initially, visual verification is skipped.
If no consensus is reached, we verify objects in the final population and merge the trajectories into a single output:
\begin{equation}
    \tau^* \sim p_\theta\big(
    \cdot \mid \mathbf{x}, \mathcal{T}_T,
    \{\mathcal{V}_j^{(T)}\}_{j=1}^{N}
    \big).
\end{equation}


\section{\textsc{Cashew-RL}~\raisebox{-1.0ex}{\includegraphics[height=1.6em]{img/cashew.png}}}
\label{sec:gspo}

To equip the model with robust multimodal aggregation capabilities, we introduce \textsc{Cashew-RL}, a two-stage post-training framework: (1) \textit{supervised fine-tuning (SFT) for trajectory aggregation}, which teaches the model to combine multiple trajectories into coherent outputs, and (2) \textit{reinforcement learning (RL) via GSPO}, which further optimizes the aggregation policy using reward signals reflecting aggregation quality. 

\subsection{Data Construction and Aggregation Format} 
Our post-training framework relies on a unified multimodal aggregation format and two datasets derived from a pool of seven image/video benchmarks (details in Appendix~\ref{app:data}). These datasets serve distinct purposes for SFT and RL stages, while sharing a consistent output structure.

\paragraph{Structured Aggregation Format.}\ 
Both SFT and RL use the output format \texttt{<think>} $r$ \texttt{</think>}, \texttt{<visual\_keys>} $K$ \texttt{</visual\_keys>}, and \texttt{<answer>} $a$ \texttt{</answer>}, where $r$ is the aggregated reasoning trace, $K$ contains relevant object-level visual evidence, and $a$ is the final answer.
This representation supports both structured supervision and RL reward computation.

\paragraph{SFT Data.}\ We first construct a 30k-instance SFT dataset containing multimodal inputs $\mathbf{x}$, questions $q$, ground-truth answers $y^\ast$, and reasoning chains generated by Qwen3-VL-30B-Thinking~\cite{bai2025qwen3vl}.
Visual keys are extracted from these chains and validated against the visual input using Grounding DINO, with manual checks on randomly sampled instances. 
These examples provide supervision for learning structured trajectory aggregation.

\paragraph{RL Data.}\ We curate a 200k-instance corpus from the same source datasets. Each instance contains the original annotation $(\mathbf{x}, q, y^\ast)$, three diverse candidate trajectories $\{\tau_i\}_{i=1}^{3}$ generated using Qwen3-VL-30B-Thinking, and a shared set of visual keys extracted from the candidates. These keys are validated against the visual input using Grounding DINO, with manual checks on randomly sampled instances. Each $\tau_i$ provides a candidate answer and a reasoning chain, which may be correct, incorrect, or incomplete.
During RL training, the candidate pool is constructed as a curriculum-based mixture of offline teacher-generated trajectories and on-policy trajectories sampled from the current model $\pi_\theta$. This hybrid setup exposes the policy to both teacher signals and noisier self-generated trajectories, reducing distribution mismatch and improving robustness. 

\paragraph{Data Split and Isolation.}\ All SFT and RL samples are constructed from publicly available splits of the respective source datasets, using official training splits where available. 
For datasets that also appear in our evaluation benchmarks, we explicitly remove any overlapping evaluation instances to prevent train-test leakage. The exact split usage and overlap filtering protocol are described in Appendix~\ref{app:data}.

\subsection{Stage I: SFT}
The goal of the supervised stage is to teach the model the \emph{structure} of multimodal aggregation: how to articulate intermediate reasoning, verbalize evidence via visual keys, and format the final answer. Given $(\mathbf{x}, q, \{\tau_i\})$, where $\{\tau_i\}$ are reference trajectories in the structured aggregation format, the model is trained to imitate a structured aggregation output $(r, K, a)$. This stage does not primarily optimize correctness; rather, it provides: (1) a stable and structured textual interface for subsequent RL training, with a well-defined output format, and (2) a reference policy $\pi_{\text{ref}}$ used for KL regularization in GSPO. Without this SFT initialization, RL training often collapses into malformed or degenerate trajectories.

\subsection{Stage II: RL via GSPO}

While SFT establishes the structural format of multimodal aggregation, it relies primarily on imitation and does not explicitly teach the model to distinguish correct evidence from hallucinated or noisy candidate trajectories. To explicitly optimize for effective aggregation, we employ reinforcement learning to train the model as an aggregator.

\paragraph{Aggregation Policy.} During RL, the model is treated as a policy
\[
\pi_\theta(a, K, r \mid \mathbf{x}, S),
\]
where $\pi_\theta$ is initialized from the reference policy $\pi_{\text{ref}}$
learned during SFT and $S = \{\tau_i\}_{i=1}^M$ denotes a set of candidate trajectories drawn from a mixture of offline cached trajectories and on-policy rollouts, following a curriculum-based teacher-on-policy trajectory mixing strategy. Each rollout consists of a reasoning trace $r$, a set of predicted visual keys $K$, and the final aggregated answer $a$.

\paragraph{Curriculum-Based Teacher-On-Policy Trajectory Mixing.} The candidate pool~$S$ evolves during RL training and is constructed as a mixture of teacher-generated trajectories and on-policy trajectories sampled from the current model~$\pi_\theta$. We employ a staged curriculum that gradually increases the proportion of on-policy candidates: early stages rely primarily on teacher trajectories to provide clear and reliable evidence, while later stages increasingly incorporate noisier and more diverse self-generated trajectories that the policy must ultimately aggregate at inference time. Appendix Table~\ref{tab:curriculum_ablation} compares this curriculum with fixed candidate compositions. The specific curriculum schedule and mixing ratios are detailed in Section~\ref{sec:experiment}.

\paragraph{Reward Design.}
Each rollout is evaluated by a composite reward that accounts for answer correctness, evidence selection quality, and adaptive reasoning efficiency:
\begin{equation}
R(a, K, r \mid \mathbf{x})
=
w_{\text{acc}}\, R_{\text{acc}}
+
w_{\text{key}}\, R_{\text{key}}
+
R_{\text{len}}.
\label{eq:reward}
\end{equation}

\noindent (1) \emph{Answer correctness ($R_{\text{acc}}$).} We use an exact-match reward $R_{\text{acc}}=\mathbf{1}[a=y^\ast]$ to provide binary, task-agnostic supervision.

\noindent (2) \emph{Evidence selection quality ($R_{\text{key}}$)} To explicitly encourage faithful visual grounding, we reward the selection of relevant visual evidence via a balanced precision-recall formulation. Let $G$ denote the ground-truth visual-key set derived from our RL training data annotations, and let $K$ be the model-predicted key set. We define:

\begin{equation}
R_{\text{key}}
=
(1-\alpha)
\frac{|K \cap G|}{|G|+\epsilon}
+
\alpha
\frac{|K \cap G|}{|K|+\epsilon}.
\end{equation}
Unlike recall-only objectives that incentivize indiscriminate key generation, this weighted formulation enables explicit control over the precision-recall trade-off. In practice, it discourages uncontrolled key proliferation while preserving sensitivity to missing critical evidence.

\noindent (3) \emph{Difficulty-Aware Length Penalty ($R_{\text{len}}$).}
Inspired by adaptive computation allocation~\cite{xiang2025justthinking}, we regulate reasoning length based on the empirical solve rate.
For each prompt, we estimate the solve rate from $J$ rollouts as $\hat p_{\text{solve}}(q) = \frac{1}{J}\sum_{j=1}^{J}\mathbf{1}[a_j=y^\ast]$, and smooth it across training steps using an exponential moving average (EMA), $\tilde p_t \leftarrow \gamma\tilde p_{t-1} + (1-\gamma)\hat p_{\text{solve}}(q)$.

Let $N_{\text{tok}}$ denote the number of tokens in the \texttt{<think>} region. 
The length penalty is defined as
\begin{equation}
R_{\text{len}} = -\beta N_{\text{tok}}\max(\tilde p_t,\,1/J).
\label{eq:alp}
\end{equation}
Prompts with higher solve rates receive stronger penalties for extended reasoning, encouraging concise aggregation while preserving flexibility for difficult prompts. 
The lower bound $1/J$ prevents the penalty from vanishing, while $\beta$ controls its strength.


Overall, this composite reward encourages the policy to produce correct answers grounded in minimal yet sufficient visual evidence, while allocating reasoning effort adaptively based on task difficulty.

\paragraph{Group Sequence Policy Optimization (GSPO).}
We optimize the aggregation policy using GSPO~\cite{zheng2025gspo}.
For each input $\mathbf{x}$ and candidate set $S$, we sample $J$ aggregation rollouts $\{y_j=(r_j,K_j,a_j)\}_{j=1}^{J}$ from the old policy $\pi_{\theta_{\mathrm{old}}}$.
Each rollout receives the composite reward $R_j$ defined in Eq.~\ref{eq:reward}.
We compute the group-relative advantage $\hat A_j$ by centering and normalizing rewards within each group.

GSPO computes a length-normalized importance ratio for
each complete aggregation rollout:
\begin{equation}
s_j(\theta)= \left(\frac{\pi_\theta(y_j\mid\mathbf{x},S)}{\pi_{\theta_{\mathrm{old}}}(y_j\mid\mathbf{x},S)}\right)^{1/|y_j|},
\label{eq:gspo_ratio}
\end{equation}
where $|y_j|$ denotes the rollout length in tokens.
For brevity, let $s_j=s_j(\theta)$ and define $\bar{s}_j=\operatorname{clip}(s_j, 1-\varepsilon_c,1+\varepsilon_c)$.
The GSPO objective with KL regularization is
\begin{equation}
\begin{aligned}
\mathcal{L}_{\mathrm{GSPO}}&=-\frac{1}{J}\sum_{j=1}^{J}\min\big(s_j\hat A_j,\bar{s}_j\hat A_j\big)\\&\quad+\alpha_{\mathrm{KL}}\widehat{\mathcal{D}}_{\mathrm{KL}}.
\end{aligned}
\label{eq:gspo}
\end{equation}
Here, $\varepsilon_c=0.2$ specifies symmetric clipping, and $\alpha_{\mathrm{KL}}=0.02$ controls regularization against the SFT reference policy $\pi_{\mathrm{ref}}$.
The KL term is estimated at the token level and averaged over the generated response.
By applying importance sampling and clipping at the sequence level, GSPO optimizes aggregation using rewards assigned to complete reasoning trajectories.

\section{Experiments}
\label{sec:experiment}
We apply \textsc{Cashew} on multiple backbone models, including InternVL-3.5~\cite{wang2025internvl35}, and Qwen3-VL~\cite{bai2025qwen3vl}. Descriptions of all evaluation benchmarks are provided in Appendix~\ref{app:benchmarks}. Baselines are evaluated using the default \texttt{LMMs-Eval} setup~\citep{zhang2025lmmseval}. For each input, we sample a population of $N{=}8$ trajectories and form aggregation groups of size $M{=}4$ (i.e., $M$ responses are grouped to synthesize one candidate); we use $T=3$ (two aggregation updates). Decoding is performed with temperature $=0.8$ and $top_p=0.95$. Visual grounding verification is provided by a frozen Grounding DINO model. All inference experiments use 8 H100 GPUs. Sensitivity to the population size $N$ and iteration number $T$ is analyzed in Appendix Figure~\ref{fig:tts_ablation_nt}.

For \textsc{Cashew-RL}, we fine-tune Qwen3-VL-4B, Qwen3-VL-8B, and InternVL3.5-8B using LoRA. RL fine-tuning is performed on 16 H100 GPUs with a global batch size of~64 and a learning rate of $1\times10^{-6}$, using the ms-swift~\cite{zhao2025swift} framework. We use sequence-level importance sampling for GSPO, with a symmetric clipping coefficient of 0.2 and a KL regularization coefficient of 0.02. For each RL instance, we construct an evidence pool of $M=4$ candidate trajectories. To balance stability and robustness, we adopt a staged curriculum that gradually increases the proportion of on-policy candidates generated by the current policy~$\pi_\theta$: $(3{:}1)$ in early training, $(2{:}2)$ in the middle stage, and $(1{:}3)$ in the final stage. This curriculum progressively exposes the model to increasingly realistic evidence distributions, with the goal of improving robustness when aggregating trajectories during inference.

\begin{table*}[t]
\centering
\scriptsize
\caption{\footnotesize\textbf{Results of \textsc{Cashew} and \textsc{Cashew-RL} on image benchmarks.} $I$: SEED-Bench results are reported only for the image subset. Improvements from \textsc{Cashew} and \textsc{Cashew-RL} are highlighted in \textcolor{darkgreen}{green} and reported as mean values. 95\% confidence intervals (CIs, ±) were computed via bootstrap resampling. Improvements are statistically significant at $\alpha = 0.05$ if the corresponding CI excludes zero. $^{\S}$ denotes non-significant improvements ($p > 0.05$; 95\% CI includes zero).}
\label{tab:image_tts}
\begin{tabular}{lcccc}
\toprule
\textbf{Model} & \textbf{ScienceQA} & \textbf{MME} & \textbf{POPE} & $\textbf{SEED-Bench}^{I}$ \\
\midrule
LLaVA-1.5-7B~\cite{liu2024llava}  & 66.8 & 302.1/1506.2 & 85.9 & 66.1 \\
Qwen-VL-Chat-7B~\cite{bai2023qwenvl}  & 68.2 & 392.1/1467.8 & 74.9 & 58.2 \\
VILA1.5-13B~\cite{lin2024vila} & 79.1 & 288.9/1429.3 & 84.2 & 62.8 \\
LLaVA-Next-7B~\cite{liu2024llavanext} & 73.0 & 308.9/1512.3 & 87.3 & 72.4 \\
LLaVA-OneVision-7B~\cite{llavaov} & 95.4 & 415.7/1577.8 & 87.4 & 75.4 \\
\midrule
InternVL3.5-8B~\cite{wang2025internvl35} & 95.9 & 663.2/1686.6 & 88.1 & 77.7 \\
\rowcolor{yellow}
\hspace{3mm}\textbf{+ \textsc{Cashew}}
& \textbf{97.8 $\scriptstyle\textcolor{darkgreen}{\textbf{(+1.9)}}$}
& \textbf{685.7/1700.2 $\scriptstyle\textcolor{darkgreen}{\textbf{(+22.5 / +13.6)}}$}
& \textbf{89.7 $\scriptstyle\textcolor{darkgreen}{\textbf{(+1.6)}}$}
& \textbf{79.0 $\scriptstyle\textcolor{darkgreen}{\textbf{(+1.3)}}$} \\
\rowcolor{blue!8}
\hspace{3mm}\textbf{+ \textsc{Cashew-RL}}
& \textbf{97.9 $\scriptstyle\textcolor{darkgreen}{\textbf{(+2.0)}}$}
& \textbf{691.8/1707.0 $\scriptstyle\textcolor{darkgreen}{\textbf{(+28.6 / +20.4)}}$}
& \textbf{90.2 $\scriptstyle\textcolor{darkgreen}{\textbf{(+2.1)}}$}
& \textbf{80.5 $\scriptstyle\textcolor{darkgreen}{\textbf{(+2.8)}}$} \\
Qwen3-VL-4B~\cite{bai2025qwen3vl} & 69.5 & 638.6/1693.7 & 88.0 & 78.7 \\
\rowcolor{yellow}
\hspace{3mm}\textbf{+ \textsc{Cashew}}
& \textbf{93.1 $\scriptstyle\textcolor{darkgreen}{\textbf{(+23.6)}}$}
& \textbf{710.4/1756.0 $\scriptstyle\textcolor{darkgreen}{\textbf{(+71.8/+62.3)}}$}
& \textbf{89.1 $\scriptstyle\textcolor{darkgreen}{\textbf{(+1.1)}}$}
& \textbf{79.8 $\scriptstyle\textcolor{darkgreen}{\textbf{(+1.1)}}$} \\
\rowcolor{blue!8}
\hspace{3mm}\textbf{+ \textsc{Cashew-RL}}
& \textbf{95.7 $\scriptstyle\textcolor{darkgreen}{\textbf{(+26.2)}}$}
& \textbf{719.7/1764.1 $\scriptstyle\textcolor{darkgreen}{\textbf{(+81.1/+70.4)}}$}
& \textbf{89.6 $\scriptstyle\textcolor{darkgreen}{\textbf{(+1.6)}}$}
& \textbf{81.2 $\scriptstyle\textcolor{darkgreen}{\textbf{(+2.5)}}$} \\
Qwen3-VL-8B~\cite{bai2025qwen3vl} & 92.9 & 643.2/1720.3 & 88.9 & 78.7 \\
\rowcolor{gray!6}
\hspace{3mm}\textit{+ \textsc{Cashew} (w/o Grounding DINO)}
& \textit{96.3 $\scriptstyle\textcolor{darkgreen}{\textit{(+3.4)}}$}
& \textit{711.0/1748.2 $\scriptstyle\textcolor{darkgreen}{\textit{(+67.8/+27.9)}}$}
& \textit{89.3 $\scriptstyle\textcolor{lightgray}{\textit{(+0.4)}}$}$^{\S}$
& \textit{79.7 $\scriptstyle\textcolor{darkgreen}{\textit{(+1.0)}}$} \\
\rowcolor{yellow}
\hspace{3mm}\textbf{+ \textsc{Cashew}}
& \textbf{97.7 $\scriptstyle\textcolor{darkgreen}{\textbf{(+4.8)}}$}
& \textbf{738.2/1772.0 $\scriptstyle\textcolor{darkgreen}{\textbf{(+95.0/+51.7)}}$}
& \textbf{89.9 $\scriptstyle\textcolor{darkgreen}{\textbf{(+1.0)}}$}
& \textbf{80.3 $\scriptstyle\textcolor{darkgreen}{\textbf{(+1.6)}}$} \\
\rowcolor{blue!8}
\hspace{3mm}\textbf{+ \textsc{Cashew-RL}}
& \textbf{97.8 $\scriptstyle\textcolor{darkgreen}{\textbf{(+4.9)}}$}
& \textbf{740.1/1769.8 $\scriptstyle\textcolor{darkgreen}{\textbf{(+96.9/+49.5)}}$}
& \textbf{90.2 $\scriptstyle\textcolor{darkgreen}{\textbf{(+1.3)}}$}
& \textbf{80.8 $\scriptstyle\textcolor{darkgreen}{\textbf{(+2.1)}}$} \\
\bottomrule
\end{tabular}
\end{table*}



\begin{table*}[t]
\centering
\caption{\footnotesize\textbf{Results of \textsc{Cashew} and \textsc{Cashew-RL} on video understanding and reasoning benchmarks.} $\dagger$ Results on Video-MME are reported without subtitles. $\ddagger$ Results on Video-TT are reported only for multiple-choice questions. Improvements from \textsc{Cashew} and \textsc{Cashew-RL} are highlighted in \textcolor{darkgreen}{green} and reported as mean values. 95\% confidence intervals (CIs, ±) were computed via bootstrap resampling. Improvements are statistically significant at $\alpha = 0.05$ if the corresponding CI excludes zero. $^{\S}$ denotes non-significant improvements ($p > 0.05$; 95\% CI includes zero).}
\label{tab:video_Cashew_all}
\resizebox{\textwidth}{!}{%
\begin{tabular}{lccccc|cccc}
\toprule
& \multicolumn{5}{c}{\textbf{Video Understanding}} 
& \multicolumn{4}{c}{\textbf{Video Reasoning}} \\
\cmidrule(lr){2-6} \cmidrule(lr){7-10}
\textbf{Model} 
& \textbf{Video-MME}$^{\dagger}$ 
& \textbf{LongVideoBench} 
& \textbf{EgoSchema} 
& \textbf{MVBench} 
& \textbf{NExT-QA}
& \textbf{VideoMMMU} 
& \textbf{VSI-Bench} 
& \textbf{Video-TT}$^{\ddagger}$ 
& \textbf{TOMATO} \\
\midrule



LLaVA-NeXT-Video-7B~\cite{zhang2024llavanextvideo} 
& -- & 43.5 & 43.9 & 46.5 & -- 
& 36.1 & 35.6 & 41.8 & 24.9 \\  

VILA1.5-40B~\cite{lin2024vila} 
& 60.1 & -- & 58.0 & -- & 67.9 
& 34.0 & 31.2 & -- & 24.7 \\

LLaVA-OneVision-7B~\cite{llavaov} 
& 58.2 & 56.4 & 60.1 & 56.7 & 79.4 
& 33.9 & 32.4 & -- & 25.5 \\ 

VideoLLaMA3-7B~\cite{zhang2025videollama3} 
& 66.2 & 59.8 & 63.3 & 69.7 & 84.5 
& 47.0 & -- & -- & -- \\

\midrule


InternVL3.5-8B~\cite{wang2025internvl35}
& 63.2 & 61.3 & 62.0 & 71.4 & 78.6 
& 50.0 & 53.2 & 43.4 & 23.8 \\

\rowcolor{yellow}
\hspace{3mm}\textbf{+ \textsc{Cashew}}
& \textbf{63.9 $\scriptstyle\textcolor{lightgray}{\textbf{(+0.7)}}$}$^{\S}$
& \textbf{62.9 $\scriptstyle\textcolor{darkgreen}{\textbf{(+1.6)}}$}
& \textbf{69.9 $\scriptstyle\textcolor{darkgreen}{\textbf{(+7.9)}}$}
& \textbf{73.0 $\scriptstyle\textcolor{darkgreen}{\textbf{(+1.6)}}$}
& \textbf{80.0 $\scriptstyle\textcolor{darkgreen}{\textbf{(+1.4)}}$}
& \textbf{50.8 $\scriptstyle\textcolor{darkgreen}{\textbf{(+0.8)}}$}
& \textbf{54.5 $\scriptstyle\textcolor{darkgreen}{\textbf{(+1.3)}}$}
& \textbf{44.2 $\scriptstyle\textcolor{lightgray}{\textbf{(+0.8)}}$}$^{\S}$
& \textbf{24.6 $\scriptstyle\textcolor{darkgreen}{\textbf{(+0.8)}}$} \\
\rowcolor{blue!8}
\hspace{3mm}\textbf{\textsc{+ Cashew-RL}}
& \textbf{64.8 $\scriptstyle\textcolor{darkgreen}{\textbf{(+1.6)}}$} 
& \textbf{63.2 $\scriptstyle\textcolor{darkgreen}{\textbf{(+1.9)}}$} 
& \textbf{71.1 $\scriptstyle\textcolor{darkgreen}{\textbf{(+9.1)}}$} 
& \textbf{73.7 $\scriptstyle\textcolor{darkgreen}{\textbf{(+2.3)}}$} 
& \textbf{80.4 $\scriptstyle\textcolor{darkgreen}{\textbf{(+1.8)}}$} 
& \textbf{51.3 $\scriptstyle\textcolor{darkgreen}{\textbf{(+1.3)}}$} 
& \textbf{54.6 $\scriptstyle\textcolor{darkgreen}{\textbf{(+1.4)}}$} 
& \textbf{44.6 $\scriptstyle\textcolor{darkgreen}{\textbf{(+1.2)}}$} 
& \textbf{24.9 $\scriptstyle\textcolor{darkgreen}{\textbf{(+1.1)}}$} \\

Qwen3-VL-4B~\cite{bai2025qwen3vl}
& 64.2 & 61.0 & 67.6 & 65.7 & 73.8 
& 46.0 & 56.6 & 40.4 & 27.6 \\

\rowcolor{yellow}
\hspace{3mm}\textbf{+ \textsc{Cashew}}
& \textbf{65.5 $\scriptstyle\textcolor{darkgreen}{\textbf{(+1.3)}}$}
& \textbf{63.6 $\scriptstyle\textcolor{darkgreen}{\textbf{(+2.6)}}$}
& \textbf{73.0 $\scriptstyle\textcolor{darkgreen}{\textbf{(+5.4)}}$}
& \textbf{68.1 $\scriptstyle\textcolor{darkgreen}{\textbf{(+2.4)}}$}
& \textbf{78.8 $\scriptstyle\textcolor{darkgreen}{\textbf{(+5.0)}}$}
& \textbf{47.2 $\scriptstyle\textcolor{darkgreen}{\textbf{(+1.2)}}$}
& \textbf{60.3 $\scriptstyle\textcolor{darkgreen}{\textbf{(+3.7)}}$}
& \textbf{41.2 $\scriptstyle\textcolor{darkgreen}{\textbf{(+0.8)}}$}
& \textbf{28.1 $\scriptstyle\textcolor{lightgray}{\textbf{(+0.5)}}$}$^{\S}$ \\
\rowcolor{blue!8}
\hspace{3mm}\textbf{\textsc{+ Cashew-RL}}
& \textbf{67.1 $\scriptstyle\textcolor{darkgreen}{\textbf{(+2.9)}}$} 
& \textbf{64.5 $\scriptstyle\textcolor{darkgreen}{\textbf{(+3.5)}}$} 
& \textbf{74.1 $\scriptstyle\textcolor{darkgreen}{\textbf{(+6.5)}}$} 
& \textbf{68.6 $\scriptstyle\textcolor{darkgreen}{\textbf{(+2.9)}}$} 
& \textbf{79.1 $\scriptstyle\textcolor{darkgreen}{\textbf{(+5.3)}}$} 
& \textbf{48.0 $\scriptstyle\textcolor{darkgreen}{\textbf{(+2.0)}}$} 
& \textbf{61.2 $\scriptstyle\textcolor{darkgreen}{\textbf{(+4.6)}}$} 
& \textbf{42.9 $\scriptstyle\textcolor{darkgreen}{\textbf{(+2.5)}}$} 
& \textbf{30.0 $\scriptstyle\textcolor{darkgreen}{\textbf{(+2.4)}}$} \\
Qwen3-VL-8B~\cite{bai2025qwen3vl} 
& 66.9 & 63.3 & 71.2 & 66.2 & 75.6 
& 47.3 & 58.8 & 43.3 & 31.5 \\
\rowcolor{gray!6}
\hspace{3mm}\textit{+ \textsc{Cashew} (w/o Grounding DINO)}
& \textit{67.5 $\scriptstyle\textcolor{lightgray}{\textit{(+0.6)}}$}$^{\S}$
& \textit{64.3 $\scriptstyle\textcolor{darkgreen}{\textit{(+1.0)}}$}
& \textit{72.5 $\scriptstyle\textcolor{darkgreen}{\textit{(+1.3)}}$}
& \textit{68.5 $\scriptstyle\textcolor{darkgreen}{\textit{(+2.3)}}$}
& \textit{80.0 $\scriptstyle\textcolor{darkgreen}{\textit{(+4.4)}}$}
& \textit{48.0 $\scriptstyle\textcolor{darkgreen}{\textit{(+0.7)}}$}
& \textit{59.5 $\scriptstyle\textcolor{darkgreen}{\textit{(+0.7)}}$}
& \textit{44.0 $\scriptstyle\textcolor{lightgray}{\textit{(+0.7)}}$}$^{\S}$
& \textit{32.9 $\scriptstyle\textcolor{darkgreen}{\textit{(+1.4)}}$} \\
\rowcolor{yellow}
\hspace{3mm}\textbf{+ \textsc{Cashew}}
& \textbf{68.3 $\scriptstyle\textcolor{darkgreen}{\textbf{(+1.4)}}$}
& \textbf{64.8 $\scriptstyle\textcolor{darkgreen}{\textbf{(+1.5)}}$}
& \textbf{74.7 $\scriptstyle\textcolor{darkgreen}{\textbf{(+3.5)}}$}
& \textbf{69.3 $\scriptstyle\textcolor{darkgreen}{\textbf{(+3.1)}}$}
& \textbf{80.5 $\scriptstyle\textcolor{darkgreen}{\textbf{(+4.9)}}$}
& \textbf{48.4 $\scriptstyle\textcolor{darkgreen}{\textbf{(+1.1)}}$}
& \textbf{61.2 $\scriptstyle\textcolor{darkgreen}{\textbf{(+2.4)}}$}
& \textbf{44.2 $\scriptstyle\textcolor{lightgray}{\textbf{(+0.9)}}$}$^{\S}$
& \textbf{33.4 $\scriptstyle\textcolor{darkgreen}{\textbf{(+1.9)}}$} \\

\rowcolor{blue!8}
\hspace{3mm}\textbf{\textsc{+ Cashew-RL}}
& \textbf{68.9 $\scriptstyle\textcolor{darkgreen}{\textbf{(+2.0)}}$} 
& \textbf{65.4 $\scriptstyle\textcolor{darkgreen}{\textbf{(+2.1)}}$} 
& \textbf{75.5 $\scriptstyle\textcolor{darkgreen}{\textbf{(+4.3)}}$} 
& \textbf{69.8 $\scriptstyle\textcolor{darkgreen}{\textbf{(+3.6)}}$} 
& \textbf{80.5 $\scriptstyle\textcolor{darkgreen}{\textbf{(+4.9)}}$} 
& \textbf{49.0 $\scriptstyle\textcolor{darkgreen}{\textbf{(+1.7)}}$} 
& \textbf{61.4 $\scriptstyle\textcolor{darkgreen}{\textbf{(+2.6)}}$} 
& \textbf{44.8 $\scriptstyle\textcolor{darkgreen}{\textbf{(+1.5)}}$} 
& \textbf{34.0 $\scriptstyle\textcolor{darkgreen}{\textbf{(+2.5)}}$} \\

\bottomrule
\end{tabular}
}
\vspace{-3mm}
\end{table*}

\subsection{Results}

\paragraph{\textsc{Cashew} Evaluation.} Across image and video benchmarks, \textsc{Cashew} consistently improves performance for diverse VLM backbones, demonstrating robustness across model scales and modalities, with most gains statistically significant under bootstrap-based confidence intervals. On image understanding tasks (Table~\ref{tab:image_tts}), \textsc{Cashew} yields substantial gains for all evaluated models. For example, Qwen3-VL-4B improves from 69.5 percentage points to 93.1 percentage points on ScienceQA, while Qwen3-VL-8B improves from 643.2/1720.3 to 738.2/1772.0 on MME. These results indicate that multi-trajectory aggregation effectively enhances both factual accuracy and perceptual grounding, regardless of backbone capacity. On video benchmarks (Table~\ref{tab:video_Cashew_all}), \textsc{Cashew} consistently improves understanding and reasoning across all tested backbones. Notably, it achieves +7.9 percentage points on EgoSchema for InternVL3.5-8B 
, highlighting its effectiveness on long-horizon tasks. Consistent gains are also observed on VSI-Bench, including +3.7 percentage points for Qwen3-VL-4B. Even on more challenging benchmarks such as Video-TT and TOMATO, \textsc{Cashew} shows positive and often statistically significant trends. Overall, these results indicate that \textsc{Cashew} strengthens temporal coherence and refines noisy reasoning trajectories, leading to more reliable and grounded video understanding and reasoning.

To understand the role of visual grounding, we ablate the Grounding DINO verification step while keeping all other aggregation mechanisms unchanged ($N{=}8$, $M{=}4$, $T{=}3$). Even without grounding, \textsc{Cashew} substantially improves over the base VLM, e.g., ScienceQA increases from 92.9 to 96.3 and NeXT-QA from 75.6 to 80.0, indicating that iterative trajectory aggregation is the primary driver of performance gains, with visual grounding further enhancing reliability and evidence alignment. Appendix Tables~\ref{tab:grounding_ablation}-\ref{tab:detection_noise} further evaluate alternative verifiers, grounding without aggregation, Self-Synthesizer with grounding, threshold sensitivity, and robustness to simulated detection noise.


\begin{table}[h]
\centering
\scriptsize
\caption{Comparison with state-of-the-art test-time scaling methods. Best scores are shown in \textbf{bold}, and second-best scores are \underline{underlined}.
}
\label{tab:sota}
\resizebox{0.48\textwidth}{!}{%
\begin{tabular}{lccccc}
\toprule
\textbf{Model} &
\textbf{ScienceQA} &
\textbf{MME} &
\textbf{EgoSchema} &
\textbf{MVBench} &
\textbf{VideoMMMU} \\
\midrule
Qwen3-VL-8B
& 92.9
& 643.2/\underline{1720.3}
& 71.2
& 66.2
& \underline{47.3} \\
\hspace{3mm}+ Self-Consistency
& 94.2
& 669.3/1702.1
& 71.1
& 66.8
& 46.4 \\
\hspace{3mm}+ Self-Selector
& 87.3
& 508.2/1431.2
& 70.1
& \underline{67.8}
& 46.7 \\
\hspace{3mm}+ Self-Synthesizer
& \underline{95.4}
& \underline{690.0}/1689.4
& \underline{72.0}
& 67.6
& 47.1 \\
\midrule
\rowcolor{yellow}
\hspace{3mm}+ \textbf{\textsc{Cashew}}
& \textbf{97.7}
& \textbf{738.2/1772.0}
& \textbf{74.7}
& \textbf{69.3}
& \textbf{48.4} \\
\bottomrule
\end{tabular}}
\vspace{-3mm}
\end{table}

\paragraph{\textsc{Cashew-RL} Evaluation.}\ We evaluate \textsc{Cashew-RL} under $T=3$ to assess whether post-training improves the learned aggregation behavior.\ Across image and video benchmarks, \textsc{Cashew-RL} consistently achieves further improvements over both the baseline and the corresponding \textsc{Cashew} results, with most gains statistically significant.\ On image understanding tasks (Table~\ref{tab:image_tts}), \textsc{Cashew-RL} yields substantial gains across backbone scales and model families. For example, on Qwen3-VL-8B, ScienceQA improves from 92.9 to 97.8 and POPE increases to 90.2.\ Importantly, similar trends are observed for the smaller Qwen3-VL-4B backbone and the non-Qwen InternVL3.5-8B backbone, indicating that the learned aggregation policy generalizes across model capacities and architectures.\ On video benchmarks (Table~\ref{tab:video_Cashew_all}), \textsc{Cashew-RL} further enhances both video understanding and reasoning.\ For Qwen3-VL-8B, EgoSchema increases from 71.2 to 75.5 and TOMATO from 31.5 to 34.0.\ Comparable improvements are observed on Qwen3-VL-4B and InternVL3.5-8B, showing consistent gains on all benchmarks over both the baseline and \textsc{Cashew}. Moreover, results in Appendix~\ref{app:iteration} show that \textsc{Cashew-RL} reaches performance comparable to higher-depth \textsc{Cashew} configurations with fewer aggregation iterations, suggesting improved aggregation efficiency. 
Appendix Tables~\ref{tab:ablation_sft_gspo}, \ref{tab:rl_disentangle}, and \ref{tab:rl_method_comparison} further disentangle the effects of SFT, generic RL, trajectory aggregation, and the choice of policy optimization algorithm.
Overall, these results indicate that post-training with GSPO strengthens the learned aggregation policy, leading to consistent improvements in multimodal reasoning across different backbone capacities and architectures.

\paragraph{Comparison with State-of-the-Art Test-Time Scaling Methods.} We compare \textsc{Cashew} with three widely used test-time scaling baselines under an $N{=}8$ multi-sample setting.\ (1) \textit{Self-Consistency}~\cite{wang2023selfconsistency} selects a final answer via majority voting over multiple sampled responses, effective when outputs are discrete and comparable. (2) \textit{Self-Selector}~\cite{parmar2025plangen} replaces majority voting with model-based judgment, using the VLM to evaluate and select a single trajectory. (3) \textit{Self-Synthesizer}~\cite{li2025reasoningaslogicunits, li2025llmsgeneratebetteranswer} goes beyond selection by generating a new response that integrates information from multiple candidate trajectories. As shown in Table~\ref{tab:sota}, \textsc{Cashew} consistently achieves the best performance across benchmarks.\ On the MME benchmark, it improves over \textit{Self-Synthesizer} by +48.2/+82.6 points on cognition and perception scores, respectively.\ On EgoSchema, \textsc{Cashew} outperforms \textit{Self-Synthesizer} by +2.7 percentage points, demonstrating the effectiveness of iterative aggregation for complex video reasoning. 
At $T=3$, \textsc{Cashew} outperforms Self-Consistency and Self-Synthesizer by 2.0 and 2.4 points on average across all 13 benchmarks at comparable VLM call counts and latency, as shown in Appendix Table~\ref{tab:compute_budget}.
Appendix Figure~\ref{fig:latency} reports the $T=1$ latency comparison.
Appendix Table~\ref{tab:recent_training_free_frameworks} further extends the comparison to recent training-free visual reasoning frameworks.

\begin{table}[h] 
\centering 
\small 
\setlength{\tabcolsep}{4.5pt} 
\caption{\footnotesize Stability ablation with repeated sampling. Acc. Std.: accuracy standard deviation across runs; Ans. Agr.: average answer agreement.} 
\label{tab:stability_analysis} 
\resizebox{\linewidth}{!}{ 
\begin{tabular}{llccccc} 
\toprule 
\textbf{Metric} & \textbf{Method} & \textbf{ScienceQA} & \textbf{SEED-Bench} & \textbf{NExT-QA} & \textbf{EgoSchema} & \textbf{VSI-Bench} \\ 
\midrule 
\multirow{4}{*}{Acc. Std. $\downarrow$} & Qwen3-VL-8B & 0.74 & 0.42 & 0.93 & 1.05 & 0.96 \\ 
& + \textsc{Cashew} ($T=1$) & 0.50 & 0.36 & 0.70 & 0.79 & 0.82 \\
& + \textsc{Cashew} ($T=2$) & 0.41 & 0.33 & 0.58 & 0.69 & 0.72 \\
&\textbf{+ \textsc{Cashew} ($T=3$)} & \textbf{0.33} & \textbf{0.31} & \textbf{0.51} & \textbf{0.64} & \textbf{0.63} \\ 
\midrule 
\multirow{4}{*}{Ans. Agr. $\uparrow$} & Qwen3-VL-8B & 94.1 & 88.3 & 83.5 & 80.9 & 77.4 \\ 
& + \textsc{Cashew} ($T=1$) & 96.0 & 89.4 & 86.7 & 84.3 & 80.1 \\
& + \textsc{Cashew} ($T=2$) & 96.9 & 90.3 & 88.1 & 85.6 & 81.5 \\
&\textbf{+ \textsc{Cashew} ($T=3$)} & \textbf{97.8} & \textbf{90.5} & \textbf{89.1} & \textbf{86.0} & \textbf{82.4} \\ 
\bottomrule 
\end{tabular}} 
\end{table}
\vspace{-3mm}

\paragraph{Effect of Iterative Aggregation on Prediction Stability.} Beyond average accuracy, we examine whether iterative aggregation directly mitigates the run-to-run inconsistency that motivates \textsc{Cashew}. We conduct five independent stochastic decoding runs under identical temperature and top-$p$ settings and report two complementary metrics: accuracy standard deviation across runs and average answer agreement. As shown in Table~\ref{tab:stability_analysis}, \textsc{Cashew} consistently reduces accuracy variation and increases answer agreement across all evaluated benchmarks. These improvements emerge with $T{=}1$ and become stronger as the number of iterations increases. At $T{=}3$, accuracy standard deviation decreases from 0.74 to 0.33 on ScienceQA and from 1.05 to 0.64 on EgoSchema, while answer agreement increases from 94.1 to 97.8 on ScienceQA and from 83.5 to 89.1 on NExT-QA. These results provide evidence that iterative trajectory aggregation improves stability while increasing average accuracy.




\section{Conclusion}
We present \textsc{Cashew}, an inference-time framework that stabilizes multimodal reasoning through iterative aggregation of candidate trajectories with visual verification, and \textsc{Cashew-RL}, a learned variant that internalizes this aggregation behavior. Using a composite reward within GSPO, \textsc{Cashew-RL} produces evidence-grounded answers while adaptively allocating reasoning effort based on task difficulty. Experiments on image and video benchmarks show that both methods improve accuracy and reasoning consistency, demonstrating the effectiveness of visually grounded iterative aggregation.

\clearpage

\newpage

\section*{Limitations}


In this work, we use a fixed aggregation budget to enable controlled comparisons across models and benchmarks, rather than optimizing the compute budget for each individual example. This setting is appropriate for studying the effect of iterative aggregation, but it may be suboptimal in resource-constrained applications where many easy examples may not require the same amount of aggregation. \textsc{Cashew-RL} partially addresses this issue by internalizing aggregation behavior and improving aggregation efficiency, but a systematic study of adaptive budget allocation remains future work.

\section*{Acknowledgments}
This research was supported by the National Eye Institute (NEI) of the National Institutes of Health (NIH) under award number R01EY034562.\ The content is solely the responsibility of the authors and does not necessarily represent the official views of the NIH. 

\bibliography{custom}

\clearpage
\appendix

\section*{Appendix}
\label{sec:appendix}

\section{\textsc{Cashew} Pseudocode}
\label{app:algo}

Algorithm~\ref{alg:Cashew} presents the pseudocode for \textsc{Cashew}, detailing its iterative aggregation and grounded object verification.

\begin{algorithm}[h]
\caption{Iterative Aggregation with Grounded Verification (\textsc{Cashew})}
\small
\label{alg:Cashew}
\begin{algorithmic}[1]
\Require Visual input $I$, question $q$, base model $p_\theta$, populations $T$, population size $N$, subset size $M$

\Function{CheckConsensus}{$P$}
    \State Let $a_1,\dots,a_{|P|}$ be the predicted answers from trajectories in $P$
    \State \Return $\bigl(a_1 = a_2 = \cdots = a_{|P|}\bigr)$
\EndFunction

\State $P_1 = \{\tau_i^{(1)} \sim p_\theta(\cdot \mid I, q)\}_{i=1}^N$

\If{\Call{CheckConsensus}{$P_1$}}
    \State \Return the common answer in $P_1$
\EndIf

\For{$t = 1$ to $T-1$}
    \For{$i = 1$ to $N$}
        \State Sample subset $S_i^{(t)} \subset P_t$ of size $M$
        \For{$\tau_j^{(t)} \in S_i^{(t)}$}
            \State Extract objects $\mathcal{O}_j^{(t)} = \mathcal{E}(r_j^{(t)})$
            \State Verify with Grounding DINO to obtain verified objects $\mathcal{V}_j^{(t)}$
        \EndFor
        \State Aggregate with grounding hints:
        \Statex \hspace{1em}$\tau_i^{(t+1)} \sim p_\theta(\cdot \mid I, q, S_i^{(t)}, \{\mathcal{V}_j^{(t)}\}_{\tau_j^{(t)}\in S_i^{(t)}})$
    \EndFor

    \State Form new population $P_{t+1} = \{\tau_i^{(t+1)}\}_{i=1}^N$

    \If{\Call{CheckConsensus}{$P_{t+1}$}}
        \State \Return the common answer in $P_{t+1}$
    \EndIf
\EndFor

\For{$\tau_j^{(T)} \in P_T$}
    \State Extract objects $\mathcal{O}_j^{(T)} = \mathcal{E}(r_j^{(T)})$
    \State Verify with Grounding DINO to obtain $\mathcal{V}_j^{(T)}$
\EndFor

\State Final aggregation:
\Statex \hspace{1em}$\tau^* \sim p_\theta(\cdot \mid I, q, P_T, \{\mathcal{V}_j^{(T)}\}_{\tau_j^{(T)}\in P_T})$
\State \Return $\tau^*$

\end{algorithmic}
\end{algorithm}

\section{\textsc{Cashew-RL} Data Generation}
\label{app:data}

\begin{table}[h]
\centering
\caption{\textsc{Cashew-RL} training dataset distribution for SFT and RL stages.}
\label{tab:dataset_distribution}
\resizebox{0.47\textwidth}{!}{%
\begin{tabular}{l|c|c}
\toprule
\textbf{Dataset} & \textbf{SFT (\# Samples)} & \textbf{RL (\# Samples)} \\
\midrule
\multicolumn{3}{c}{\textbf{Image Datasets}} \\
\midrule
GQA~\cite{hudson2019gqa} & 8{,}300 & 76{,}000 \\
MathVista~\cite{lu2024mathvista} & 400 & 5{,}000 \\
ScienceQA~\cite{lu2022scienceqa} & 2{,}000 & 10{,}000 \\
\midrule
\multicolumn{3}{c}{\textbf{Video Datasets}} \\
\midrule
SSV2~\cite{goyal2017ssv2} & 8{,}500 & 40{,}000 \\
NExT-QA~\cite{xiao2021nextqa} & 600 & 2{,}500 \\
Ego4D~\cite{grauman2022ego4d} & 9{,}800 & 65{,}000 \\
EgoTextVQA~\cite{zhou2025egotextvqa} & 400 & 1{,}500 \\
\midrule
\textbf{Total} & \textbf{30{,}000} & \textbf{200{,}000} \\
\bottomrule
\end{tabular}}
\end{table}

As summarized in Table~\ref{tab:dataset_distribution}, we construct the \textsc{Cashew-RL} training set from a diverse collection of image and video datasets to expose the model to a wide range of visual inputs and reasoning scenarios. The image datasets include general visual question answering, mathematical reasoning, and science-oriented tasks, while the video datasets cover short-term actions, long-horizon temporal reasoning, and egocentric understanding. This diverse composition encourages robust, generalizable aggregation behaviors rather than specialization to a single task or modality. Table~\ref{tab:dataset_distribution} also reports the data distribution for the SFT stage (30k instances) and the RL stage (200k instances).

All SFT and RL samples are constructed from publicly available splits of the respective source datasets, using official training splits where available. We remove training instances that overlap with evaluation benchmarks using exact image or video identifier matching. Concretely, for ScienceQA, we exclude all image IDs appearing in the evaluation split; for NExT-QA, we remove all video IDs used in evaluation; and for Ego4D, we exclude any video IDs that appear in EgoSchema. This filtering is performed prior to trajectory generation and visual key construction, ensuring that no evaluation samples or associated visual content are exposed during post-training.

\begin{table}[t]
\centering
\small
\setlength{\tabcolsep}{4pt}
\caption{Human verification of visual keys. Supported and relevant keys report the percentages judged visually supported and relevant to the question, respectively.
Agreement reports the percentage of identical judgments from two annotators.}
\label{tab:visual_key_audit}
\begin{tabular}{lccc}
\toprule
\textbf{Metric} & \textbf{SFT} & \textbf{RL} & \textbf{Overall} \\
\midrule
Instances           & 1,000  & 1,000  & 2,000  \\
\midrule
Supported Keys      & 94.1\% & 92.8\% & 93.4\% \\
Relevant Keys       & 89.3\% & 87.5\% & 88.4\% \\
Presence Agreement  & 96.1\% & 94.2\% & 95.2\% \\
Relevance Agreement & 90.7\% & 93.3\% & 92.1\% \\
\bottomrule
\end{tabular}
\vspace{-3mm}
\end{table}

\paragraph{Human verification of visual keys.} The teacher proposes candidate visual keys, and Grounding DINO checks their presence in the visual input.
To assess the quality of these targets, we manually inspected 2,000 randomly sampled training instances, including 1,000 from the SFT set and 1,000 from the RL set.
Four graduate annotators each reviewed 1,000 instances, with every instance independently reviewed by two annotators.
For each visual key, annotators judged whether it was visually supported and relevant to answering the question.
Disagreements were resolved by a third annotator.
Table~\ref{tab:visual_key_audit} reports the proportions of supported and relevant keys, together with raw agreement between annotators.
Agreement is computed as the percentage of keys for which the two annotators gave the same judgment.
The audit shows that most sampled visual keys are supported by the visual input and relevant to the question, with high agreement between annotators on both criteria.




\section{Additional Implementation Details}

Table~\ref{tab:hyperparameters} summarizes the hyperparameter configurations used for inference and training. We implement the object parser $\mathcal{E}$ using a \texttt{spaCy} noun phrase parser that extracts compound object mentions and removes pronouns and meta terms.

\begin{table}[h]
\centering
\caption{Complete Hyperparameter Configuration.}
\label{tab:hyperparameters}
\scriptsize
\begin{tabular}{l|c|c}
\toprule
\textbf{Parameter Description} & \textbf{Symbol} & \textbf{Value}\\ 
\midrule
\multicolumn{3}{l}{\textsc{Cashew}/\textsc{Cashew-RL}: \textit{\textbf{Inference Phase}}} \\ 
\midrule
Candidate population size per iteration & $N$ & 8\\
Iteration steps & $T$ & 3\\
Candidates aggregated per group & $M$ & 4\\
Grounding DINO confidence threshold & $\delta_g$ & 0.35\\
\midrule
\multicolumn{3}{l}{{\textsc{Cashew-RL}: \textit{\textbf{Training Phase}}}} \\ 
\midrule
Global batch size & $B$ & 64\\
Learning rate & $\eta$ & $1 \times 10^{-6}$\\
Number of rollouts per prompt & $J$ & 4\\ 
\midrule
\multicolumn{3}{l}{{\textsc{Cashew-RL}: \textit{\textbf{Reward Function Design (Eq.7 - Eq.9)}}}} \\ 
\midrule
Weight for answer accuracy & $w_{\text{acc}}$ & 1.0\\
Weight for visual evidence overlap & $w_{\text{key}}$ & 0.35 \\
Precision-recall balancing coefficient & $\alpha$ & 0.5 \\
Epsilon for division stability & $\epsilon$ & $1\text{e-}8$\\
Adaptive length penalty coefficient & $\beta$ & 0.001 \\
EMA decay for solve rate & $\gamma$ & 0.9\\ 
\midrule
\multicolumn{3}{l}{{\textsc{Cashew-RL}:
\textit{\textbf{GSPO (Eq.10 - Eq.11)}}}} \\
\midrule
Importance sampling level & -- & Sequence \\
Clipping coefficient & $\varepsilon_c$ & 0.2 \\
Upper clipping override & \texttt{epsilon\_high} & None \\
KL regularization coefficient & $\alpha_{\mathrm{KL}}$ & 0.02 \\

\bottomrule
\end{tabular}%
\vspace{-3mm}
\end{table}

\paragraph{Artifact availability.}
We will release the code, checkpoints, split identifiers,
preprocessing and data generation scripts, and derived annotations, subject to the licenses of the source datasets.

\begin{table*}[h]
\centering
\scriptsize
\caption{Study on verifier robustness in \textsc{Cashew} under Qwen3-VL-8B.}
\label{tab:grounding_ablation}
\resizebox{0.95\textwidth}{!}{%
\begin{tabular}{lcccccc}
\toprule
\textbf{Model} & \textbf{Verifier} &  \textbf{ScienceQA} & \textbf{SEED-Bench} & \textbf{NExT-QA} & \textbf{EgoSchema} & \textbf{VSI-Bench} \\
\midrule
Qwen3-VL-8B & -- & 92.9 & 78.7 & 75.6 & 71.2 & 58.8 \\
\hspace{3mm}+ \textsc{Cashew} & VLM self-check & 96.7 & 79.7 & 80.2 & 73.2  & 60.0\\
\hspace{3mm}+ \textsc{Cashew} & GLIP & 97.5 & \textbf{80.5} & \textbf{80.6} & 74.3  & 61.0\\
\rowcolor{yellow}
\hspace{3mm}+ \textsc{Cashew} & Grounding DINO 
& \textbf{97.7}
& 80.3 
& 80.5 
& \textbf{74.7} 
& \textbf{61.2} \\
\bottomrule
\end{tabular}}
\end{table*}

\begin{table*}[h]
\centering
\small
\caption{Study on grounding without trajectory aggregation in \textsc{Cashew}.}
\label{tab:grounding_no_aggregation}
\begin{tabular}{lcccccc}
\toprule
\textbf{Model} & \textbf{ScienceQA} & \textbf{SEED-Bench} & \textbf{NExT-QA} & \textbf{EgoSchema} & \textbf{VSI-Bench} \\
\midrule
Qwen3-VL-8B  & 92.9 & 78.7 & 75.6 & 71.2 & 58.8 \\
\hspace{3mm}+ Grounding DINO & 93.2 & 78.3 & 76.0 & 71.6 & 58.2 \\
\midrule
\hspace{3mm}+ \textsc{Cashew} ($T=1$) & 94.6 & 78.9 & 76.8 & 72.4 & 58.6 \\
\rowcolor{yellow}
\hspace{3mm}\textbf{+ \textsc{Cashew}} ($T=3$) & \textbf{97.7} \textbf{$\scriptstyle\textcolor{darkgreen}{\textbf{(+4.8)}}$} 
& \textbf{80.3} \textbf{$\scriptstyle\textcolor{darkgreen}{\textbf{(+1.6)}}$} 
& \textbf{80.5} \textbf{$\scriptstyle\textcolor{darkgreen}{\textbf{(+4.9)}}$} 
& \textbf{74.7} \textbf{$\scriptstyle\textcolor{darkgreen}{\textbf{(+3.5)}}$} 
& \textbf{61.2} \textbf{$\scriptstyle\textcolor{darkgreen}{\textbf{(+2.4)}}$} \\
\bottomrule
\end{tabular}
\vspace{-2mm}
\end{table*}

\section{Ablation Studies}
\label{app:ablation}

\paragraph{RQ1: What is the role of visual grounding verification in \textsc{Cashew}?} 
\label{app:visual_grounding}
We analyze visual grounding verification from three perspectives: (1) replacing the Grounding DINO with other verifiers, (2) introducing grounding without aggregation, and (3) adding grounding to Self-Synthesizer.

\noindent(1) \emph{Effect of replacing Grounding DINO}. 
We further replace Grounding DINO with alternative verification signals, including VLM self-check and GLIP~\citep{li2022glip}. Both variants generally improve over the no-grounding setting, showing that \textsc{Cashew} benefits from visual verification signals independent of the specific verifier. Grounding DINO yields the strongest overall results, further boosting performance from 96.3 to 97.7 on ScienceQA and from 72.5 to 74.7 on EgoSchema. These results suggest that grounding acts as a stabilizing mechanism that reduces hallucinations. However, the additional gains from grounding are smaller than those from aggregation itself, confirming that aggregation remains the dominant driver of improvement.

\noindent(2) \emph{Effect of grounding without aggregation}. To further disentangle the effect of grounding from aggregation, we evaluate a single-rollout variant with grounding verification but without trajectory aggregation (i.e., no population sampling or iterative synthesis). Table~\ref{tab:grounding_no_aggregation} shows that incorporating grounding alone provides only marginal improvements over the base VLM (e.g., +0.3 percentage points on ScienceQA and +0.4 percentage points on EgoSchema), whereas multi-trajectory aggregation results in much larger gains (+4.8 percentage points and +3.5 percentage points, respectively). This confirms that the performance gains arise primarily from the verification-guided aggregation mechanism, not from the detector.

\begin{table}[t]
\centering
\scriptsize
\setlength{\tabcolsep}{3pt}
\caption{Adding Grounding DINO to Self-Synthesizer on Qwen3-VL-8B.
All variants use $N=8$ initial trajectories; \textsc{Cashew} uses $T=3$.}
\label{tab:synthesizer_grounding}
\resizebox{\linewidth}{!}{%
\begin{tabular}{lcccc}
\toprule
\textbf{Method} &
\textbf{ScienceQA} &
\textbf{EgoSchema} &
\textbf{VSI-Bench} &
\textbf{TOMATO} \\
\midrule
Self-Synthesizer
& 95.4 & 72.0 & 59.1 & 32.0 \\
\hspace{3mm}+ Grounding DINO
& 95.6 & 72.6 & 59.2 & 32.2 \\
\rowcolor{yellow}
\textbf{\textsc{Cashew}}
& \textbf{97.7} & \textbf{74.7} & \textbf{61.2} & \textbf{33.4} \\
\bottomrule
\end{tabular}}
\end{table}

\noindent(3) \emph{Adding grounding to Self-Synthesizer.} We equip Self-Synthesizer with the same object verification signal used in \textsc{Cashew}, while retaining its single synthesis step. All variants use Qwen3-VL-8B and $N=8$ initial trajectories. Table~\ref{tab:synthesizer_grounding} shows that adding Grounding DINO yields modest gains for Self-Synthesizer, while \textsc{Cashew} outperforms this grounded variant by 2.1 points on EgoSchema and 2.0 points on VSI-Bench. Furthermore, across all 13 benchmarks, adding Grounding DINO raises Self-Synthesizer's average score from 67.4 to 67.7, while \textsc{Cashew} reaches 69.4. These results indicate that adding visual verification to a single synthesis step is insufficient to match the performance of the complete \textsc{Cashew} framework.

\paragraph{RQ2: How robust is \textsc{Cashew} to the visual grounding verification?} We examine how sensitive \textsc{Cashew} is to the design and quality of visual grounding verification from two complementary perspectives: (1) varying the grounding confidence threshold, and (2) perturbing detection outputs with simulated noise.

\begin{table}[h]
\centering
\scriptsize
\caption{Study on the sensitivity of \textsc{Cashew} and \textsc{Cashew-RL} to the grounding confidence threshold $\delta_g$.}
\label{tab:threshold_sensitivity}
\begin{tabular}{lcccc}
\toprule
\textbf{Model} & $\delta_g$ & \textbf{ScienceQA} & \textbf{EgoSchema} \\
\midrule
Qwen3-VL-8B & -- & 92.9 & 71.2 \\
\midrule
\hspace{3mm}+ \textsc{Cashew} & 0.25 & 97.0 & 74.0 \\
\hspace{3mm}+ \textsc{Cashew} & 0.30 & 97.4 & 74.5 \\
\rowcolor{yellow}
\hspace{3mm}\textbf{+ \textsc{Cashew}} & 0.35 & \textbf{97.7} & \textbf{74.7} \\
\hspace{3mm}+ \textsc{Cashew} & 0.45 & 97.3 & 74.4 \\
\midrule
\hspace{3mm}+ \textsc{Cashew-RL} & 0.25 & 97.5 & 75.2 \\
\hspace{3mm}+ \textsc{Cashew-RL} & 0.30 & 97.7 & 75.3 \\
\rowcolor{blue!8}
\hspace{3mm}\textbf{+ \textsc{Cashew-RL}} & 0.35 & \textbf{97.8} & \textbf{75.5} \\
\hspace{3mm}+ \textsc{Cashew-RL} & 0.45 & 97.6 & 75.3 \\
\bottomrule
\end{tabular}
\vspace{-3mm}
\end{table}

\noindent(1) \emph{Sensitivity to Grounding Confidence Threshold}. If \textsc{Cashew} heavily relied on precise detector calibration, small changes in the grounding confidence threshold $\delta_g$ would cause large performance fluctuations. To test this, we vary $\delta_g \in \{0.25, 0.30, 0.35, 0.45\}$ on ScienceQA and EgoSchema. In all main experiments, we use $\delta_g = 0.35$ as a default setting. As shown in Table~\ref{tab:threshold_sensitivity}, performance remains stable across practical thresholds. For \textsc{Cashew}, variations are within $\pm$0.7 percentage points on ScienceQA and EgoSchema. \textsc{Cashew-RL} shows even smaller fluctuations (within $\pm$0.3 percentage points). Only extreme thresholds result in moderate degradation, which is expected when verification becomes overly permissive or overly restrictive. These results show that \textsc{Cashew} is not hypersensitive to grounding confidence tuning, and does not rely on delicate detector calibration.

\begin{table}
\centering
\scriptsize
\caption{Study on the robustness of \textsc{Cashew} and \textsc{Cashew-RL} under simulated detection noise.}
\label{tab:detection_noise}
\resizebox{0.48\textwidth}{!}{%
\begin{tabular}{llcc}
\toprule
\textbf{Model} & \textbf{Noise Type} & \textbf{ScienceQA} & \textbf{EgoSchema} \\
\midrule
Qwen3-VL-8B & No Noise & 92.9 & 71.2 \\
\midrule
\hspace{3mm}+ \textsc{Cashew} & Remove 10\% Objects & 96.8 & 73.4 \\
\hspace{3mm}+ \textsc{Cashew} & Add 10\% Objects & 96.9 & 73.6 \\
\hspace{3mm}+ \textsc{Cashew} & Remove 10\% + Add 10\% & 96.2 & 72.9 \\
\rowcolor{yellow}
\hspace{3mm}\textbf{+ \textsc{Cashew}} & No Noise & \textbf{97.7} & \textbf{74.7} \\
\midrule
\hspace{3mm}+ \textsc{Cashew-RL} & Remove 10\% Objects & 97.1 & 74.6 \\
\hspace{3mm}+ \textsc{Cashew-RL} & Add 10\% Objects & 97.2 & 74.8 \\
\hspace{3mm}+ \textsc{Cashew-RL} & Remove 10\% + Add 10\% & 96.7 & 74.1 \\
\rowcolor{blue!8}
\hspace{3mm}\textbf{+ \textsc{Cashew-RL}} & No Noise & \textbf{97.8} & \textbf{75.5} \\
\bottomrule
\end{tabular}}
\vspace{-3mm}
\end{table}

\noindent (2) \emph{Robustness to Detection Noise}. To assess dependency on perception quality, we simulate realistic detector noise by randomly removing 10\% of verified objects (false negatives) and injecting 10\% additional low-confidence objects (false positives). As shown in Table~\ref{tab:detection_noise}, \textsc{Cashew} degrades only slightly (approximately 1.0 percentage points), while \textsc{Cashew-RL} shows even smaller degradation (approximately 0.8 percentage points). These results indicate that aggregation integrates grounding cues as soft evidence rather than rigid constraints, and does not critically depend on precise detector outputs.


\paragraph{RQ3: How do SFT and RL contribute to aggregation behavior?} \textsc{Cashew-RL} is trained via a two-stage post-training pipeline, consisting of supervised fine-tuning (SFT) followed by reinforcement learning with GSPO. We analyze the training stages, trajectory aggregation, and candidate composition during RL.

\begin{table*}[h]
\centering
\scriptsize
\caption{Stage-wise ablation study of supervised fine-tuning (SFT) and GSPO-based reinforcement learning (RL). SFT + RL corresponds to CASHEW-RL with a single aggregation step ($T=1$).}
\vspace{-1mm}
\label{tab:ablation_sft_gspo}
\resizebox{\textwidth}{!}{%
\begin{tabular}{lcccccc}
\toprule
\textbf{Model} & \textbf{ScienceQA} & \textbf{Video-MME} & \textbf{NExT-QA} & \textbf{EgoSchema} & \textbf{VideoMMMU} & \textbf{VSI-Bench} \\
\midrule
Qwen3-VL-8B & 92.9 & 66.9 & 75.6 & 71.2 & 47.3 & 58.8 \\
\hspace{3mm}+ SFT & 93.6 & 66.5 & 76.1 & 70.8 & 46.2 & 58.1 \\
\rowcolor{blue!8}
\hspace{3mm}+ \textbf{SFT + RL (Ours)} 
& \textbf{96.9} \textbf{$\scriptstyle\textcolor{darkgreen}{\textbf{(+4.0)}}$}
& \textbf{67.8} \textbf{$\scriptstyle\textcolor{darkgreen}{\textbf{(+0.9)}}$}
& \textbf{78.6} \textbf{$\scriptstyle\textcolor{darkgreen}{\textbf{(+3.0)}}$}
& \textbf{74.6} \textbf{$\scriptstyle\textcolor{darkgreen}{\textbf{(+3.4)}}$}
& \textbf{47.9} \textbf{$\scriptstyle\textcolor{darkgreen}{\textbf{(+0.6)}}$}
& \textbf{60.0} \textbf{$\scriptstyle\textcolor{darkgreen}{\textbf{(+1.2)}}$} \\
\bottomrule
\end{tabular}}
\end{table*}

\begin{table*}[h]
\centering
\scriptsize
\setlength{\tabcolsep}{2pt}
\caption{Ablation of generic RL and trajectory aggregation. 
Vanilla GSPO is trained without aggregation conditioning, while single rollout evaluates \textsc{Cashew-RL} without inference-time aggregation.}
\vspace{-1mm}
\label{tab:rl_disentangle}
\resizebox{\textwidth}{!}{%
\begin{tabular}{lcccccc}
\toprule
\textbf{Model} & \textbf{ScienceQA} & \textbf{Video-MME} & \textbf{NExT-QA} & \textbf{EgoSchema} & \textbf{VideoMMMU} & \textbf{VSI-Bench} \\
\midrule
Qwen3-VL-8B
& 92.9 & 66.9 & 75.6 & 71.2 & 47.3 & 58.8 \\

\hspace{3mm}+ Vanilla GSPO
& 94.4 & 67.3 & 77.2 & 72.2 & 47.4 & 59.1 \\

\hspace{3mm}+ \textsc{Cashew-RL} (single rollout)
& 93.4 & 67.1 & 76.9 & 71.4 & 47.4 & 58.8 \\

\rowcolor{blue!8}
\hspace{3mm}\textbf{+ \textsc{Cashew-RL}}
& \textbf{96.9} \textbf{$\scriptstyle\textcolor{darkgreen}{\textbf{(+4.0)}}$}
& \textbf{67.8} \textbf{$\scriptstyle\textcolor{darkgreen}{\textbf{(+0.9)}}$}
& \textbf{78.6} \textbf{$\scriptstyle\textcolor{darkgreen}{\textbf{(+3.0)}}$}
& \textbf{74.6} \textbf{$\scriptstyle\textcolor{darkgreen}{\textbf{(+3.4)}}$}
& \textbf{47.9} \textbf{$\scriptstyle\textcolor{darkgreen}{\textbf{(+0.6)}}$}
& \textbf{60.0} \textbf{$\scriptstyle\textcolor{darkgreen}{\textbf{(+1.2)}}$} \\
\bottomrule
\end{tabular}}
\vspace{-2mm}
\end{table*}

\noindent (1) \emph{Stage-wise contribution of SFT and RL}. We first evaluate three variants: (1) the base VLM without post-training, (2) SFT-only, and (3) SFT + RL (i.e., \textsc{Cashew-RL}) with a single aggregation ($T=1$). 

Results in Table~\ref{tab:ablation_sft_gspo} show that SFT alone yields mixed effects across benchmarks. While it improves performance on ScienceQA and NExT-QA, it leads to slight regressions on reasoning-oriented benchmarks such as EgoSchema and VSI-Bench. This suggests that SFT primarily enforces structural consistency in aggregation outputs but does not reliably improve evidence selection or reasoning quality.
In contrast, adding GSPO-based RL improves performance over SFT across all six benchmarks. For example, VideoMMMU improves by 1.7 points, VSI-Bench by 1.9 points, and EgoSchema by 3.8 points, all relative to SFT.
These results demonstrate the contribution of RL to improving aggregation performance beyond SFT.

\noindent (2) \emph{Disentangling RL and aggregation effects.} While the stage-wise analysis shows that RL improves performance when aggregation is enabled, it remains unclear whether these gains stem from generic RL, improved intrinsic quality of individual trajectories, or more effective aggregation behavior. 
To examine these factors, we additionally evaluate a Vanilla GSPO baseline trained with the same backbone, training data, LoRA configuration, and optimization budget, but without candidate trajectories or aggregation conditioning. 
We also evaluate \textsc{Cashew-RL} under a single-rollout setting without inference-time aggregation, which measures standalone reasoning quality after aggregation training.

Results in Table~\ref{tab:rl_disentangle} show that Vanilla GSPO yields modest gains but remains consistently below full \textsc{Cashew-RL} under the evaluated inference settings.
Similarly, \textsc{Cashew-RL} under single-rollout inference yields only limited improvements over the baseline model. 
In contrast, \textsc{Cashew-RL} with aggregation ($T=1$) achieves substantially larger gains over the base VLM, including +4.0 percentage points on ScienceQA and +3.4 percentage points on EgoSchema. 
These findings suggest that aggregation remains the dominant source of performance gains, while RL further improves performance when aggregation is enabled.

\begin{table}[t]
\centering
\scriptsize
\setlength{\tabcolsep}{3pt}
\caption{Effect of candidate composition during RL.
Models after post-training are evaluated with $T=1$.}
\label{tab:curriculum_ablation}
\resizebox{\linewidth}{!}{%
\begin{tabular}{lcccc}
\toprule
\textbf{Setting} &
\textbf{ScienceQA} &
\textbf{EgoSchema} &
\textbf{NExT-QA} &
\textbf{VSI-Bench} \\
\midrule
Qwen3-VL-8B
& 92.9 & 71.2 & 75.6 & 58.8 \\
\hspace{3mm}+ SFT
& 93.6 & 70.8 & 76.1 & 58.1 \\
\midrule
Teacher trajectories only
& 95.8 & 73.2 & 77.3 & 58.8 \\
On-policy trajectories only
& 95.1 & 72.9 & 76.7 & 58.7 \\
Fixed 1:1 mixture
& 96.4 & 73.9 & 78.0 & 59.5 \\
\rowcolor{blue!8}
\textbf{Staged curriculum}
& \textbf{96.9}
& \textbf{74.6}
& \textbf{78.6}
& \textbf{60.0} \\
\bottomrule
\end{tabular}}
\vspace{-3mm}
\end{table}

\noindent(3) \emph{Effect of candidate composition during RL.} We compare training with teacher trajectories only, on-policy trajectories only, a fixed 1:1 mixture, and the staged curriculum $3{:}1 \rightarrow 2{:}2 \rightarrow 1{:}3$. All other settings are held fixed.

Table~\ref{tab:curriculum_ablation} shows that the fixed 1:1 mixture outperforms training with either source alone, while the staged curriculum achieves the highest scores across all four benchmarks.
Compared with the fixed mixture, the curriculum improves EgoSchema by 0.7 points and NExT-QA by 0.6 points.
These results support combining teacher and on-policy candidates and gradually adjusting their proportions during training.

\paragraph{RQ4: How does reinforcement learning affect the iterative behavior of \textsc{Cashew}?}
\label{app:iteration}

\begin{figure*}[h]
    \centering
    \includegraphics[width=0.75\linewidth]{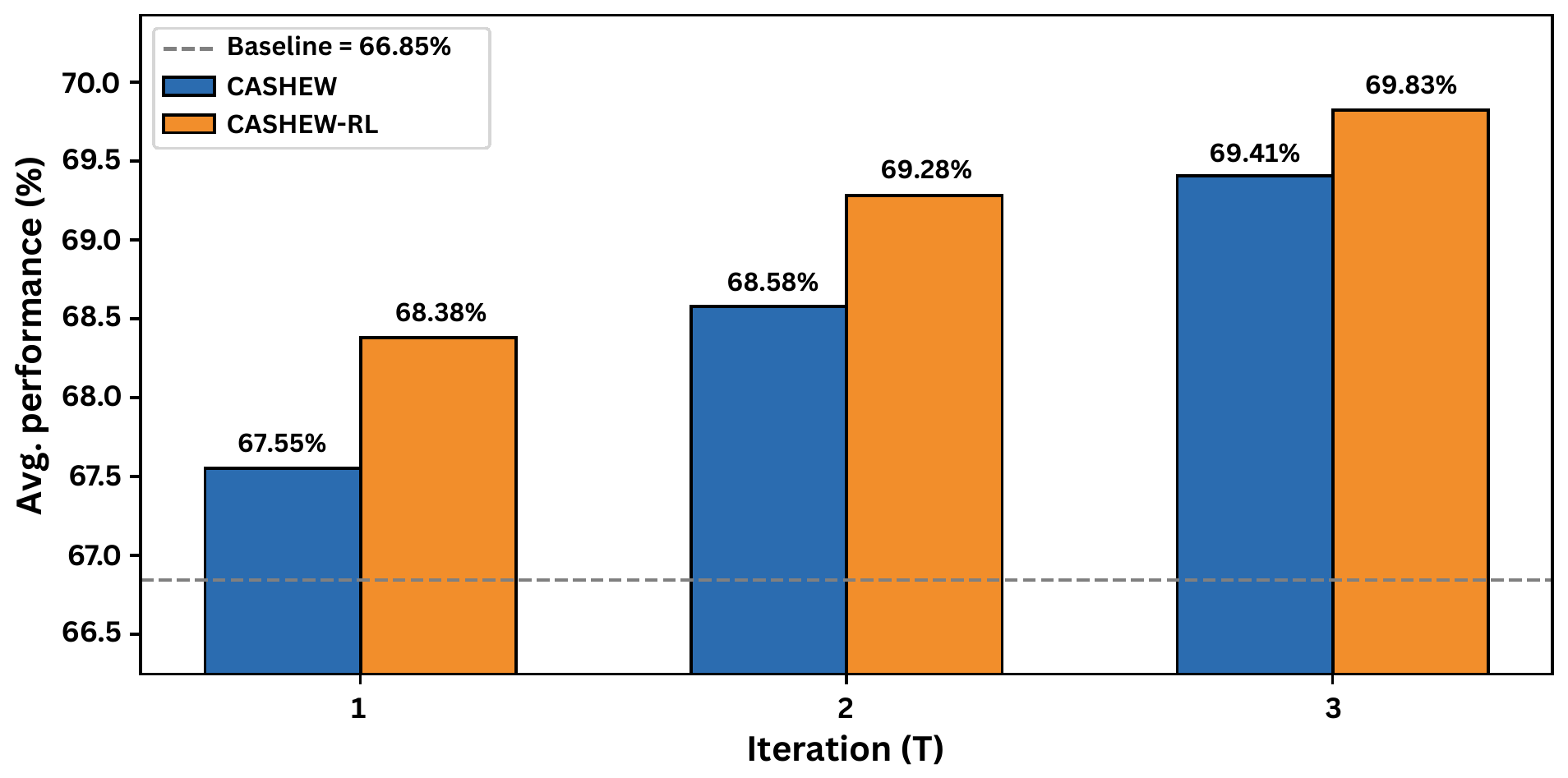}
    \vspace{-0.3cm}
    \caption{Performance comparison between \textsc{Cashew} and \textsc{Cashew-RL} across different $T$, with fixed population size $N=8$ and group size $M=4$. Performance is averaged over 13 benchmarks. For MME, scores are normalized to percentages for consistency.}
    \label{fig:rl_ablation_t}
    \vspace{-2mm}
\end{figure*}

\begin{figure*}[h]
    \centering
    \includegraphics[width=0.7\linewidth]{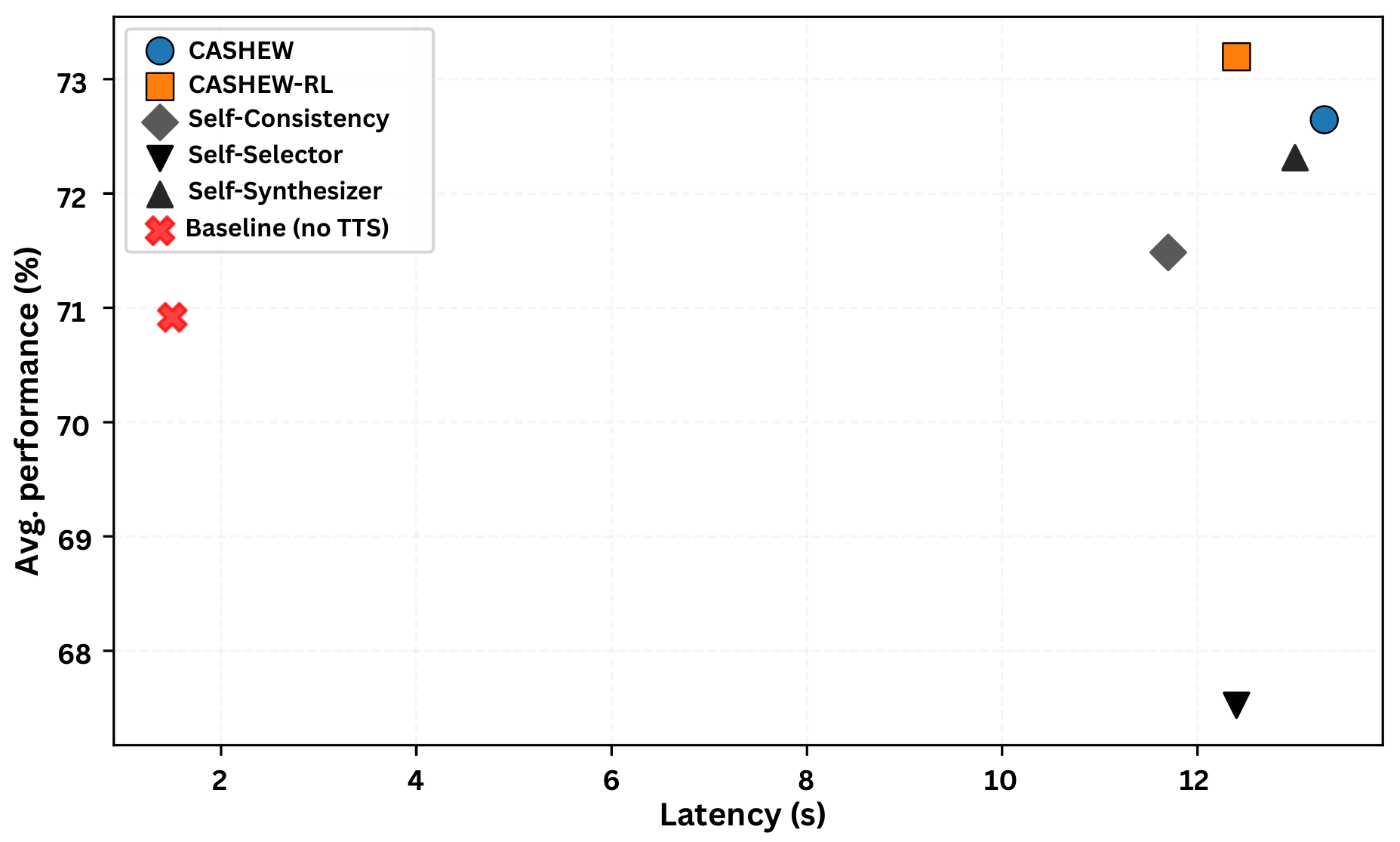}
    \caption{Latency-performance comparison between \textsc{Cashew}, \textsc{Cashew-RL}, and baselines TTS methods. \textsc{Cashew} and \textsc{Cashew-RL} are evaluated with $T=1$. Performance is averaged over the five benchmarks reported in Table~3 in the main paper, and latency is measured as end-to-end inference time per sample.}
    \label{fig:latency}
    \vspace{-2mm}
\end{figure*}

We compare \textsc{Cashew} and \textsc{Cashew-RL} across iteration numbers $T \in \{1,2,3\}$ on Qwen3-VL-8B under identical decoding settings to examine how RL training alters the aggregation trajectory. Figure~\ref{fig:rl_ablation_t} shows that \textsc{Cashew-RL} consistently outperforms \textsc{Cashew} at every iteration. While both models benefit from increasing the iteration number $T$, \textsc{Cashew-RL} starts from a stronger initial performance and maintains a higher improvement throughout the iterative aggregation process. Notably, its performance at iteration $T$ is often comparable to or exceeds that of \textsc{Cashew} at iteration $T+1$, indicating that reinforcement learning training effectively improves the aggregation behavior. 

Meanwhile, increasing $T$ improve performance for \textsc{Cashew-RL}, suggesting that RL does not eliminate the benefit of iterative aggregation. Instead, RL strengthens individual rollouts and enhances their compatibility with aggregation. Overall, RL reshapes the iterative behavior of \textsc{Cashew} by improving per-iteration effectiveness while preserving the benefits of multi-step aggregation.

\paragraph{RQ5: How does \textsc{Cashew}/\textsc{Cashew-RL} compare to existing test-time scaling methods in the latency-performance tradeoff?}
\label{app:tradeoff}

We compare \textsc{Cashew} and \textsc{Cashew-RL} against three representative TTS baselines: Self-Consistency, Self-Selector, and Self-Synthesizer, measuring both average performance and end-to-end latency. All methods are evaluated on Qwen3-VL-8B under identical configurations ($N=8$). For a fair comparison, \textsc{Cashew} and \textsc{Cashew-RL} are evaluated with a single aggregation iteration ($T=1$). As shown in Figure~\ref{fig:latency}, both \textsc{Cashew} and \textsc{Cashew-RL} achieve higher average performance than all baselines while maintaining comparable latency. Notably, \textsc{Cashew-RL} reaches the highest accuracy with even lower latency than the other TTS methods. Importantly, as shown in Figure~\ref{fig:rl_ablation_t}, increasing the iteration number $T$ further improves the performance of both \textsc{Cashew} and \textsc{Cashew-RL} when additional time budget is available. This demonstrates that \textsc{Cashew} offers a favorable latency-performance tradeoff at $T=1$, while remaining scalable under larger inference budgets.

\begin{figure*}[t]
    \centering
    \includegraphics[width=0.95\linewidth]{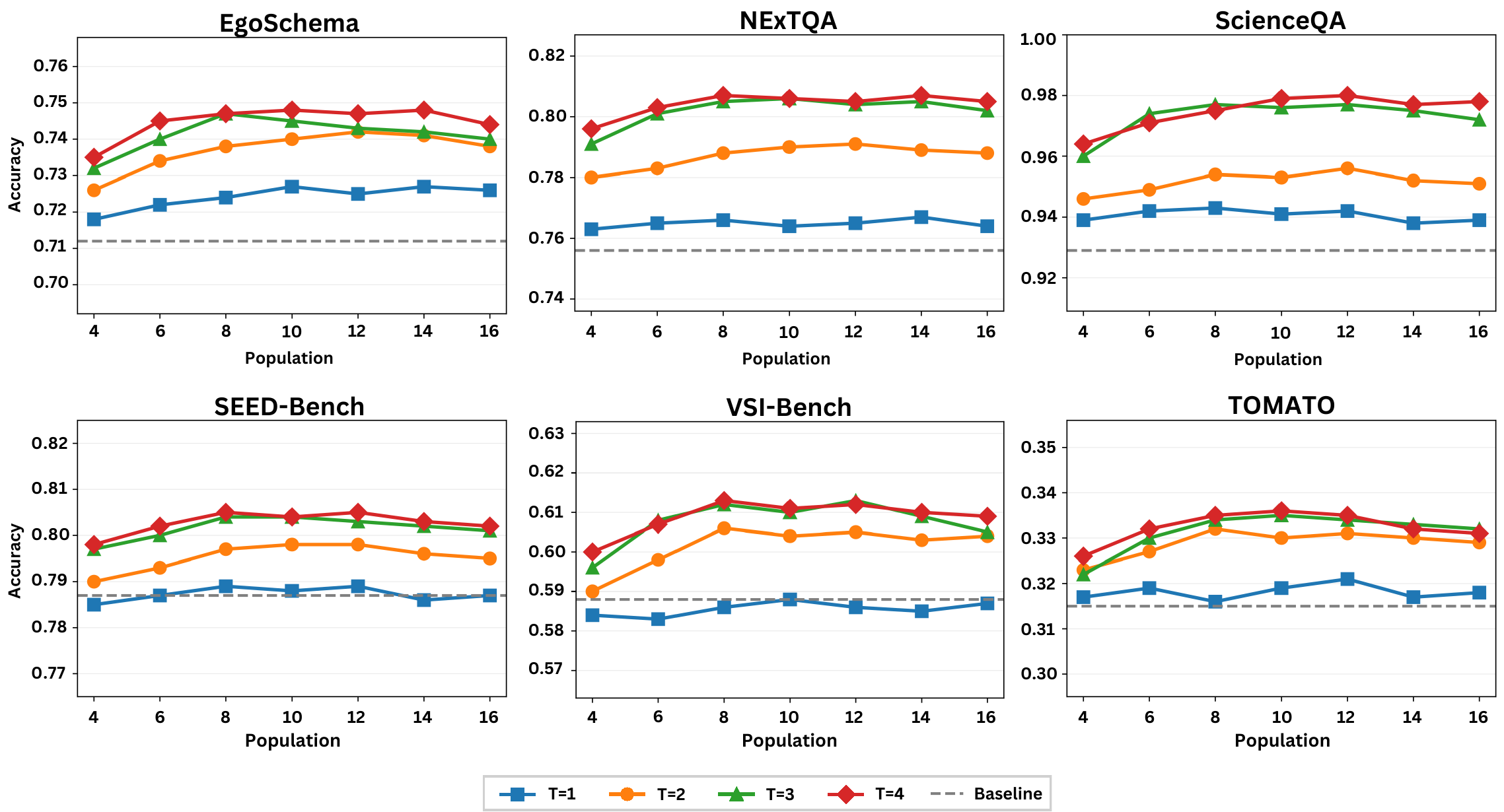}
    \caption{Performance across \textsc{Cashew} population ($N$) for different values of $T$. All results with fixed $M=4$.}
    \label{fig:tts_ablation_nt}
    \vspace{-2mm}
\end{figure*}

\begin{table}[t]
\centering
\scriptsize
\setlength{\tabcolsep}{5pt}
\caption{Comparison at comparable inference budgets.
Results summarize all 13 benchmarks.
VLM calls and latency are reported as per-example averages.}
\label{tab:compute_budget}
\resizebox{0.48\textwidth}{!}{%
\begin{tabular}{lcccc}
\toprule
\textbf{Method} &
\textbf{Setting} &
\textbf{VLM Calls} &
\textbf{Latency (s)} &
\textbf{Avg. Score} \\
\midrule
Qwen3-VL-8B
& Single pass & 1.0 & 1.5 & 66.9 \\
\midrule
Self-Consistency
& $N=8$ & 8.0 & 11.7 & 67.1 \\
Self-Selector
& $N=8$ & 9.0 & 12.4 & 66.6 \\
Self-Synthesizer
& $N=8$ & 9.0 & 13.0 & 67.4 \\
\rowcolor{yellow}
\textbf{\textsc{Cashew}}
& $N=8,\ T=1$ & 9.0 & 13.3 & \textbf{67.6} \\
\midrule
Self-Consistency
& $N=20$ & 20.0 & 29.2 & 67.4 \\
Self-Synthesizer
& $N=18$ & 19.0 & 29.8 & 67.0 \\
\rowcolor{yellow}
\textbf{\textsc{Cashew}}
& $N=8,\ T=3$ & 19.5 & 28.3 & \textbf{69.4} \\
\bottomrule
\end{tabular}}
\vspace{-3mm}
\end{table}

\noindent\textbf{Comparison at comparable inference budgets.} We additionally evaluate whether the gains persist when competing methods receive larger sampling budgets. Table~\ref{tab:compute_budget} reports average VLM call counts, end-to-end latency, and performance across all 13 benchmarks. All methods in this comparison use the same Qwen3-VL-8B checkpoint, vLLM implementation, hardware, batching configuration, and decoding settings. We count each generated initial candidate or aggregation output as one VLM call, accounting for early stopping as specified in Algorithm~\ref{alg:Cashew}. For $N=8$ and $T=3$, the maximum is 25 VLM calls, while early stopping reduces the average to 19.5. End-to-end latency includes the aggregation and visual verification overhead of \textsc{Cashew}. At $T=1$, \textsc{Cashew} outperforms Self-Synthesizer by 0.2 points with the same average VLM call count and comparable latency. At $T=3$, it exceeds Self-Consistency with $N=20$ and Self-Synthesizer with $N=18$ by 2.0 and 2.4 points, respectively. These results show that the gains persist even when competing methods receive larger sampling budgets.

\noindent\textbf{Component-wise latency.}
Table~\ref{tab:latency_breakdown} decomposes the end-to-end latency of \textsc{Cashew} into initial trajectory generation, object parsing, Grounding DINO verification, and aggregation generation. Most runtime is spent on VLM generation and aggregation. At $T=3$, object parsing and visual verification take 0.20 and 1.74 seconds, respectively, within a total latency of 28.34 seconds. Increasing $T$ primarily increases the aggregation cost.

\begin{table}[t]
\centering
\scriptsize
\setlength{\tabcolsep}{4pt}
\caption{Average per-example latency breakdown of \textsc{Cashew} on Qwen3-VL-8B. All values are in seconds.}
\label{tab:latency_breakdown}
\begin{tabular}{lcc}
\toprule
\textbf{Component} &
\textbf{$T=1$} &
\textbf{$T=3$} \\
\midrule
Initial trajectory generation & 11.42 & 11.42 \\
Object parsing & 0.05 & 0.20 \\
Grounding DINO verification & 0.56 & 1.74 \\
Aggregation generation & 1.30 & 14.98 \\
\midrule
\textbf{Total} & \textbf{13.33} & \textbf{28.34} \\
\bottomrule
\end{tabular}
\vspace{-3mm}
\end{table}

\begin{table}[h]
\centering
\scriptsize
\setlength{\tabcolsep}{4.0pt}
\caption{Comparison with recent training-free visual reasoning frameworks. MME: percentage scores.}
\label{tab:recent_training_free_frameworks}
\resizebox{0.48\textwidth}{!}{%
\begin{tabular}{l|cccccc}
\toprule
\textbf{Model} & \textbf{MME} & \textbf{MMMU} & \textbf{MathVista} & \textbf{HallusionBench} & \textbf{POPE} & \textbf{SEED-Bench} \\
\midrule
GPT-4o-mini
& 79.2 & 48.4 & 53.0 & 56.0 & 81.2 & 80.9\\
\quad + ReAct
& 87.3 & 54.8 & 49.3 & 51.1 & -- & -- \\
\quad + \textsc{ProReason}
& 91.9 & \textbf{61.6} & 54.9 & 59.9 & -- & -- \\
\rowcolor{yellow}
\quad + \textsc{Cashew}
& \textbf{92.6} & 61.4 & \textbf{57.2} & \textbf{61.4} & \textbf{83.9} & \textbf{82.7} \\
Qwen3-VL-8B
& 84.4 & 62.4 & 77.2 & 61.1 & 88.9 & 78.7\\
\quad + SAP
& 87.3 & 62.3 & -- & -- & \textbf{89.9} & 77.3 \\
\rowcolor{yellow}
\quad + \textsc{Cashew}
& \textbf{89.6} & \textbf{62.8} & \textbf{77.6} & \textbf{63.7} & \textbf{89.9} & \textbf{80.3} \\
\bottomrule
\end{tabular}}
\end{table}

\begin{table*}[t]
\centering
\scriptsize
\caption{\textbf{Comparison of GSPO and GRPO for \textsc{Cashew-RL} on Qwen3-VL-8B.} $I$: SEED-Bench results are reported only for the image subset.}
\label{tab:rl_method_comparison}
\begin{tabular}{lcccccc}
\toprule
\textbf{Model} 
& \textbf{ScienceQA} 
& $\textbf{SEED-Bench}^{I}$ 
& \textbf{EgoSchema} 
& \textbf{MVBench} 
& \textbf{VideoMMMU} 
& \textbf{VSI-Bench} \\
\midrule
Qwen3-VL-8B & 92.9 & 78.7 & 71.2 & 66.2 & 47.3 & 58.8 \\
\hspace{3mm}\textbf{\textsc{+ Cashew-RL} (GRPO)} & 97.4 & 80.1 & \textbf{75.5} & 69.3 & 48.4 & 61.1 \\
\rowcolor{blue!8}
\hspace{3mm}\textbf{\textsc{+ Cashew-RL} (GSPO)} 
& \textbf{97.8} 
& \textbf{80.8} 
& \textbf{75.5} 
& \textbf{69.8}
& \textbf{49.0} 
& \textbf{61.4} \\
\bottomrule
\end{tabular}
\vspace{-2mm}
\end{table*}

\paragraph{RQ6: How does \textsc{Cashew} compare with recent training-free visual reasoning frameworks?}\ In addition to the canonical TTS baselines compared in the main paper, we further compare \textsc{Cashew} with recent training-free visual reasoning frameworks.\ Specifically, we consider ReAct~\citep{yao2023react}, which interleaves reasoning traces with action-style steps; \textsc{ProReason}~\citep{zhou2025proreason}, which decomposes multimodal reasoning into proactive visual perception and textual reasoning; and SAP~\citep{shi2026saliency}, which performs saliency-guided multi-route inference.\ 
As shown in Table~\ref{tab:recent_training_free_frameworks}, \textsc{Cashew} achieves strong performance across both backbone settings.\ 
Under GPT-4o-mini, \textsc{Cashew} outperforms ReAct and \textsc{ProReason} on 3 out of the 4 commonly reported benchmarks.\ 
Compared with the strongest recent baseline, \textsc{ProReason}, \textsc{Cashew} improves MME from 91.9 to 92.6, MathVista~\citep{lu2024mathvista} from 54.9 to 57.2, and HallusionBench~\citep{guan2024hallusionbench} from 59.9 to 61.4. Under Qwen3-VL-8B, \textsc{Cashew} matches or outperforms SAP on all shared benchmarks. For example, \textsc{Cashew} improves MME from 87.3 to 89.6, MMMU~\citep{yue2024mmmu} from 62.3 to 62.8, and SEED-Bench from 77.3 to 80.3.\ 
These results show that \textsc{Cashew}'s iterative trajectory aggregation remains effective not only against canonical TTS baselines, but also against recent training-free visual reasoning frameworks.

\paragraph{RQ7: How do the population size $N$ and iteration number $T$ affect \textsc{Cashew}'s performance?} We study the effect of key aggregation hyperparameters in \textsc{Cashew} by varying the population size $N \in \{4,6,8,10,12,14,16\}$ and the iteration number $T \in \{1,2,3,4\}$, while fixing the group size to $M=4$. Figure~\ref{fig:tts_ablation_nt} reports results across six benchmarks, with all other decoding parameters held constant. Across tasks, increasing the population size $N$ from small to moderate values (e.g., $4 \rightarrow 8$) yields consistent gains for $T = 2, 3, 4$, while further increases provide only marginal improvements or lead to slight regressions. Performance also improves as $T$ increases from $1$ to $3$, with $T=3$ offering a strong and stable operating point across benchmarks. Increasing to $T=4$ can yield marginal gains, primarily at the cost of increased computational overhead and latency. Overall, the results indicate that a moderate population size of $N=8$ combined with $T=3$ iterations offers a favorable trade-off between performance and computational cost.


\paragraph{RQ8: How robust is \textsc{Cashew-RL} across different RL training algorithms?}

In the main paper, \textsc{Cashew-RL} is trained using GSPO, which has been reported to show stable optimization behavior in prior work~\cite{zheng2025gspo}. To evaluate whether \textsc{Cashew-RL} depends on a specific RL algorithm, we additionally train \textsc{Cashew-RL} using GRPO under identical settings, and compare it with the GSPO variant on benchmarks across different tasks. Table~\ref{tab:rl_method_comparison} shows that GSPO achieves slightly higher performance on most benchmarks, although the differences are small. For example, GSPO outperforms GRPO by 0.7 on SEED-Bench and 0.6 on VideoMMMU, while performance is nearly identical on EgoSchema. Overall, these results indicate that the gains of \textsc{Cashew-RL} are not tied to a particular RL optimizer. The aggregation-aware training objective remains effective across different policy optimization methods.

\begin{table}[t]
\centering
\scriptsize
\setlength{\tabcolsep}{3pt}
\caption{Effect of chain-of-thought reasoning in \textsc{Cashew} under identical aggregation settings.}
\vspace{-2mm}
\label{tab:cot_ablation}
\resizebox{0.48\textwidth}{!}{%
\begin{tabular}{lcccc}
\toprule
\textbf{Model} & \textbf{ScienceQA} & \textbf{EgoSchema} & \textbf{VideoMMMU} & \textbf{VSI-Bench} \\
\midrule
Qwen3-VL-8B & 92.9 & 71.2 & 47.3 & 58.8 \\
\hspace{3mm}+ \textsc{Cashew} (w/o CoT) & 96.6 & 72.3 & 47.6 & 59.7 \\
\rowcolor{yellow}
\hspace{3mm}\textbf{+ \textsc{Cashew}} & \textbf{97.7} & \textbf{74.7} & \textbf{48.4} & \textbf{61.2} \\
\bottomrule
\end{tabular}}
\vspace{-2mm}
\end{table}

\paragraph{RQ9: What is the contribution of chain-of-thought (CoT) in \textsc{Cashew}?} To isolate the contribution of explicit chain-of-thought reasoning, we compare \textsc{Cashew} with and without CoT under identical settings. As shown in Table~\ref{tab:cot_ablation}, \textsc{Cashew} without CoT already yields substantial gains over the base model (e.g., 92.9 → 96.6 on ScienceQA), indicating that iterative aggregation with visual verification is the primary source of improvement. Adding CoT further improves performance, particularly on reasoning-oriented benchmarks such as VideoMMMU and VSI-Bench, suggesting that explicit step-by-step reasoning is especially beneficial for temporal and multi-step inference tasks. Overall, aggregation remains the dominant driver of performance gains, while CoT provides additional benefits on tasks requiring structured reasoning.

\section{Evaluation Benchmarks}
\label{app:benchmarks}

We evaluate \textsc{Cashew} across three categories of multimodal benchmarks: (1) image understanding, (2) video understanding, and (3) video reasoning. Image understanding benchmarks include ScienceQA~\cite{lu2022scienceqa}, MME~\cite{fu2024mme}, POPE~\cite{li2023pope}, and SEED-Bench (image subset)~\cite{li2024seed}. Video understanding benchmarks cover both short- and long-form video comprehension, including Video-MME~\cite{fu2025videomme}, LongVideoBench~\cite{wu2024longvideobench}, EgoSchema~\cite{mangalam2023egoschema}, MVBench~\cite{li2024mvbench}, and NExT-QA~\cite{xiao2021nextqa}. Finally, video reasoning benchmarks consist of VideoMMMU~\cite{hu2025videommmu}, VSI-Bench~\cite{yang2025vsibench}, Video-TT~\cite{zhang2025videott}, and TOMATO~\cite{shangguan2025tomato}.

\section{Prompt Templates}
\label{app:prompt}

Figure~\ref{fig:Cashew_prompt} shows the prompt templates used in \textsc{Cashew}. The \textbf{population initialization prompt} generates initial candidate trajectories from the input media (image or video) and query.\ The \textbf{grounded aggregation prompt} guides iterative aggregation by incorporating verified visual objects, while the \textbf{final aggregation prompt} synthesizes information from a larger candidate set to produce a single final answer. All prompts enforce a consistent output structure, ensuring reliable aggregation.

Figure~\ref{fig:Cashew_rl_prompt} shows the prompt template used for supervised fine-tuning (SFT). The SFT aggregation prompt trains the model to consolidate multiple candidate trajectories into a single grounded answer by producing a reasoning chain, identifying relevant visual objects, and generating a concise final response. This prompt enforces the same structured output format as the test-time aggregation prompts, providing consistent supervision for learning aggregation and visual grounding behaviors.

For the test-time scaling baselines, Self-Selector uses a selection prompt that asks the model to choose one existing candidate. The instruction is: ``Select the single best response based on its correctness and consistency with the visual input. Do not rewrite or combine the responses. Return only the integer index of the best response, from 1 to N.'' This prompt permits selection but not revision, making candidate ranking a potential source of error: the model may choose an incorrect response even when a correct candidate is available. We verified that the regression in Table~\ref{tab:sota} persists across repeated runs and is not caused by output parsing failures.

In contrast, Self-Synthesizer uses an aggregation prompt to generate a new response from the candidates. Figure~\ref{fig:self_synthesizer_prompt} shows its prompt templates. Unlike \textsc{Cashew}, Self-Synthesizer performs a single round of aggregation without explicit grounding signals. The population initialization prompt is shared with \textsc{Cashew},
with only the aggregation prompt differing in structure and the information it receives.

\begin{figure*}[t]
\centering

\begin{minipage}{\textwidth}
\begin{tcolorbox}[
  colback=gray!2,
  colframe=black!75,
  title=Prompt for Population Initialization,
  fonttitle=\bfseries,
]
You are given a \texttt{\{media\_type\}} and a question.
\vspace{4pt}

\textbf{Question}:  
\texttt{\{question\}}
\vspace{4pt}

Carefully reason step by step inside \texttt{<think>...</think>} tag, then output only one concise final answer inside \texttt{<answer>...</answer>} tag.
\end{tcolorbox}
\end{minipage}

\vspace{3pt}

\begin{minipage}{\textwidth}
\begin{tcolorbox}[
  colback=gray!2,
  colframe=black!75,
  title=Prompt for Grounded Aggregation,
  fonttitle=\bfseries,
]
You are given a \texttt{\{media\_type\}}, a question, a group of candidate answers, and some objects that are already verified and important to answer the question. The candidates may be wrong or incomplete. Review them carefully and generate a new answer.
\vspace{4pt}

\textbf{Question}:  
\texttt{\{question\}}

\vspace{4pt}
\textbf{Candidate Answers}:

\#\#\# Candidate \#1 \#\#\#: \texttt{\{candidate1\}}

\#\#\# Candidate \#2 \#\#\#: \texttt{\{candidate2\}}

\#\#\# Candidate \#3 \#\#\#: \texttt{\{candidate3\}}

\#\#\# Candidate \#4 \#\#\#: \texttt{\{candidate4\}}
\vspace{4pt}

\textbf{Key Objects}:
[\texttt{\{visual\_keys\}}]
\vspace{4pt}

Write your reasoning inside \texttt{<think>...</think>} tag and end with exactly one concise answer inside \texttt{<answer>...</answer>} tag. 
\end{tcolorbox}
\end{minipage}

\vspace{3pt}

\begin{minipage}{\textwidth}
\begin{tcolorbox}[
  colback=gray!2,
  colframe=black!75,
  title=Prompt for Final Aggregation,
  fonttitle=\bfseries,
]
This is the final aggregation round. You are given a \texttt{\{media\_type\}}, a group of candidate answers, and some objects that are already verified and important to answer the question. Read \textbf{all} candidate answers carefully and aggregate useful information from them to produce \textbf{exactly one} final answer.

\textbf{Question}:  
\texttt{\{question\}}
\vspace{4pt}

\textbf{Candidate Answers}:

\#\#\# Candidate \#1 \#\#\#: \texttt{\{candidate1\}}

\#\#\# Candidate \#2 \#\#\#: \texttt{\{candidate2\}}


...


\#\#\# Candidate \#7 \#\#\#: \texttt{\{candidate7\}}

\#\#\# Candidate \#8 \#\#\#: \texttt{\{candidate8\}}
\vspace{4pt}

\textbf{Key Objects}:
[\texttt{\{visual\_keys\}}]
\vspace{4pt}

Even if the information from the candidates and the visual input is insufficient, please make your best possible guess based on the question. Write your reasoning inside \texttt{<think>...</think>} tag and end with \textbf{exactly one} final answer and place it inside \texttt{<answer>...</answer>} tag.
\end{tcolorbox}
\end{minipage}

\caption{Prompt templates for different stages of \textsc{Cashew}, including population initialization, grounded aggregation, and final aggregation.}
\label{fig:Cashew_prompt}
\end{figure*}

\begin{figure*}[t]
\centering

\begin{minipage}{\textwidth}
\begin{tcolorbox}[
  colback=gray!2,
  colframe=black!75,
  title=Prompt for supervised fine-tuning (SFT),
  fonttitle=\bfseries,
]
You are given a \texttt{\{media\_type\}}, a question, and a group of candidate answers. The candidates may be wrong or incomplete. Aggregate the useful ideas and produce a single, high-quality answer. Be concise and correct.
\vspace{4pt}

\textbf{Question}:  
\texttt{\{question\}}

\vspace{4pt}
\textbf{Candidate Answers}:

\#\#\# Candidate \#1 \#\#\#: \texttt{\{candidate1\}}

\#\#\# Candidate \#2 \#\#\#: \texttt{\{candidate2\}}

\#\#\# Candidate \#3 \#\#\#: \texttt{\{candidate3\}}

\#\#\# Candidate \#4 \#\#\#: \texttt{\{candidate4\}}
\vspace{4pt}

\textbf{Output Requirements:}

1. In <think></think>, write a single coherent reasoning chain that compares and aggregates the candidate answers, and discards incorrect or unsupported claims.

2. In <visual\_keys></visual\_keys>, output a Python-style list of objects from the visual input that are most relevant for answering the question. Only include objects that provide useful visual evidence.

3. In <answer></answer>, provide one concise and correct final answer.
\vspace{4pt}

\textbf{Output Format:}
\vspace{4pt}

<think>

Your reasoning chain here.

</think>
\vspace{4pt}

<visual\_keys>

["object\_1", "object\_2", ...]

</visual\_keys>
\vspace{4pt}

<answer>

Your final answer here.

</answer>

\end{tcolorbox}
\end{minipage}
\caption{Prompt used for supervised fine-tuning (SFT) in \textsc{Cashew-RL}. The prompt enforces a structured output format consisting of a reasoning chain, a list of visual keys, and a final answer, enabling the model to learn aggregation and grounding behaviors from demonstrations.}
\label{fig:Cashew_rl_prompt}
\end{figure*}

\begin{figure*}[t]
\centering

\begin{minipage}{\textwidth}
\begin{tcolorbox}[
  colback=gray!2,
  colframe=black!75,
  title=Prompt for Population Initialization,
  fonttitle=\bfseries,
]
You are given a \texttt{\{media\_type\}} and a question.
\vspace{3pt}

Question:  
\texttt{\{question\}}
\vspace{3pt}

Carefully reason step by step inside \texttt{<think>...</think>} tag, then output only one concise final answer inside \texttt{<answer>...</answer>} tag.
\end{tcolorbox}
\end{minipage}

\vspace{6pt}

\begin{minipage}{\textwidth}
\begin{tcolorbox}[
  colback=gray!2,
  colframe=black!75,
  title=Prompt for Self-Synthesizer Aggregation,
  fonttitle=\bfseries,
]
This is the final aggregation round. You are given a \texttt{\{media\_type\}}, and several candidate answers. Read \textbf{all} candidate answers carefully and aggregate useful information from them to produce \textbf{exactly one} final answer.

Question:  
\texttt{\{question\}}
\vspace{3pt}

Candidate Answers:

\#\#\# Candidate \#1 \#\#\#: \texttt{\{candidate1\}}

\#\#\# Candidate \#2 \#\#\#: \texttt{\{candidate2\}}

\#\#\# Candidate \#3 \#\#\#: \texttt{\{candidate3\}}

\#\#\# Candidate \#4 \#\#\#: \texttt{\{candidate4\}}

\#\#\# Candidate \#5 \#\#\#: \texttt{\{candidate5\}}

\#\#\# Candidate \#6 \#\#\#: \texttt{\{candidate6\}}

\#\#\# Candidate \#7 \#\#\#: \texttt{\{candidate7\}}

\#\#\# Candidate \#8 \#\#\#: \texttt{\{candidate8\}}
\end{tcolorbox}
\end{minipage}

\caption{Prompt templates for Self-Synthesizer in ablation study.}
\label{fig:self_synthesizer_prompt}
\end{figure*}

\section{Qualitative Examples}

\begin{figure*}[t]
 \centering
 \includegraphics[width=\linewidth]{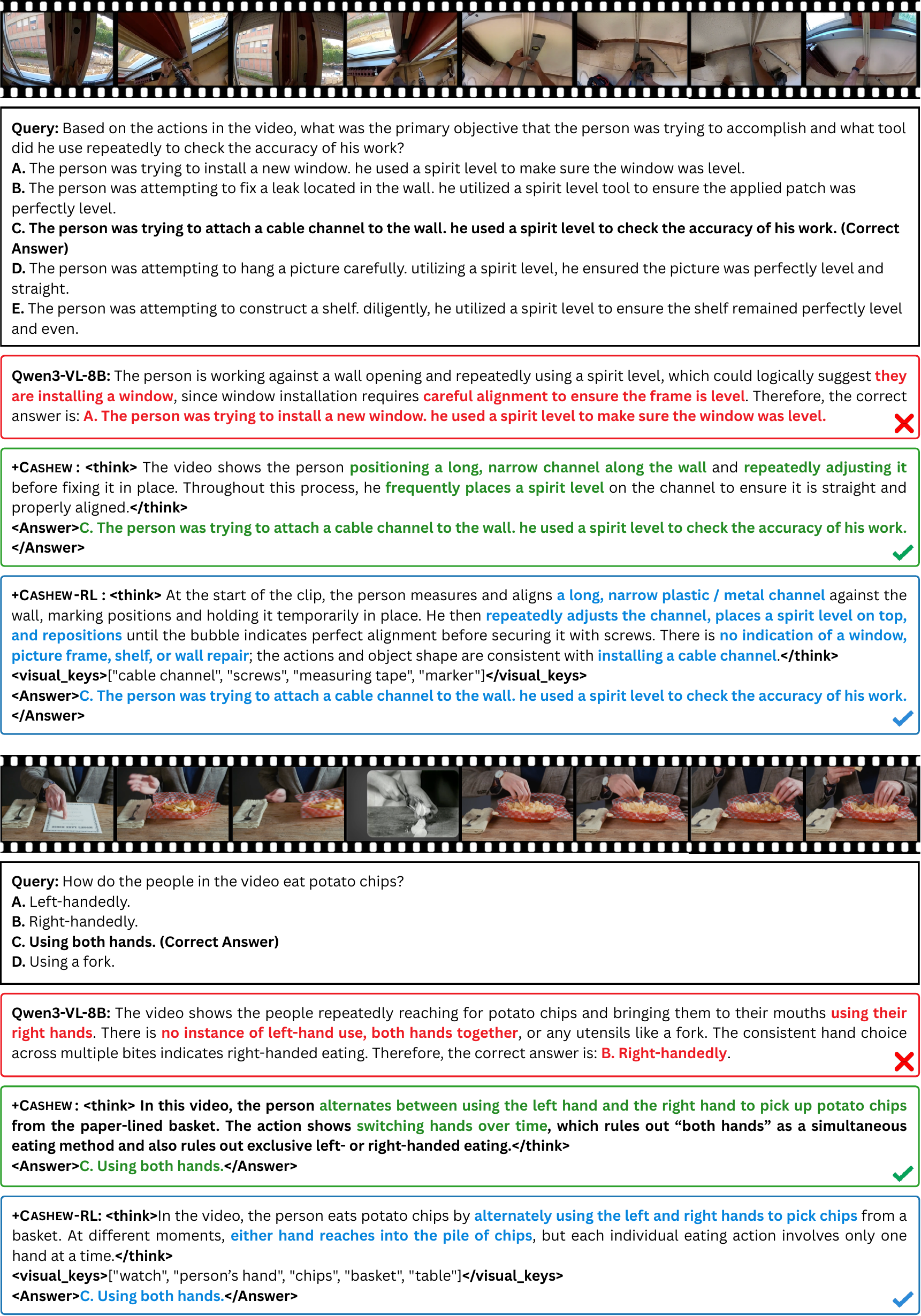}
 \caption{\textbf{Qualitative examples of \textsc{Cashew} and \textsc{Cashew-RL}}. \textbf{\textcolor{AlertRed}{Red}} text indicates errors or hallucinations produced by the baseline Qwen3-VL-8B. \textbf{\textcolor{SuccessGreen}{Green}} and \textbf{\textcolor{AzureBlue}{blue}} text highlight correct, visually grounded reasoning generated by \textsc{Cashew} and \textsc{Cashew-RL}, respectively.}
\label{fig:video_examples}
\end{figure*}

\begin{figure*}[t]
 \centering
 \includegraphics[width=0.97\linewidth]{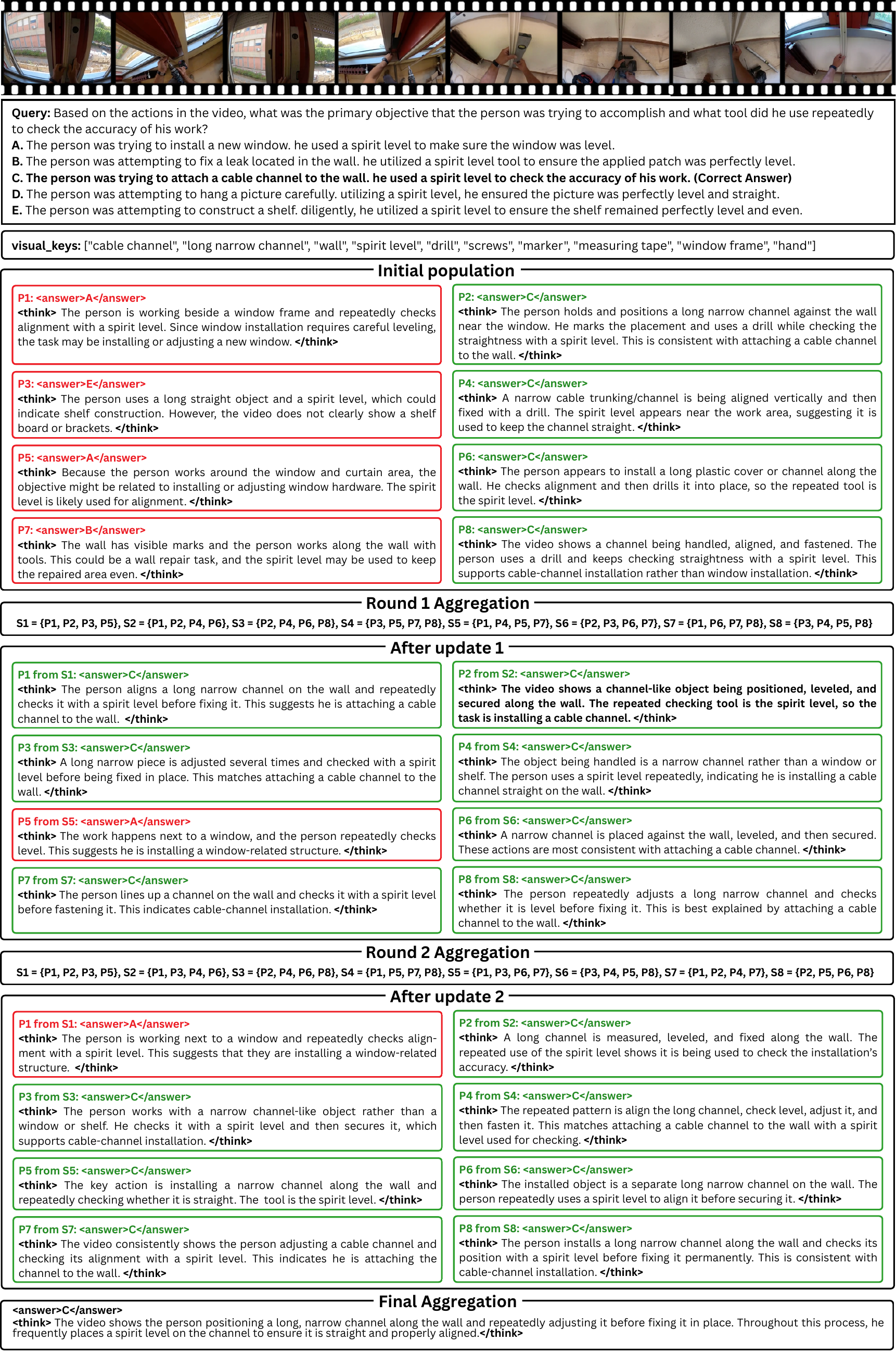}
 \caption{\textbf{Example correction process of \textsc{Cashew}.} Starting from eight trajectories with mixed predictions, \textsc{Cashew} iteratively aggregates random subsets of four trajectories with verified visual keys.}
\label{fig:process_examples}
\end{figure*}

Figure~\ref{fig:video_examples} compares the Qwen3-VL-8B baseline~\cite{bai2025qwen3vl} with \textsc{Cashew} and \textsc{Cashew-RL}. \textbf{In the first example (top)}, the baseline observes ``careful alignment to ensure the frame is level'' but over-generalizes, incorrectly assuming window installation. This reflects a factual hallucination from single-path reasoning. \textsc{Cashew} corrects this by iteratively aggregating multiple trajectories, using verified objects as grounding cues while the VLM interprets observations such as ``a long, narrow channel'', ``repeatedly adjusting it'', and ``frequently places a spirit level on the channel'', rejecting the unsupported hypotheses. \textsc{Cashew-RL} further internalizes this behavior, identifying informative visual cues like the cable channel, screws, and marker, and producing a structured, grounded conclusion. \textbf{In the second example (bottom)}, the Qwen3-VL-8B baseline fails temporally, focusing on right-hand actions despite the video showing alternating left- and right-hand use. \textsc{Cashew} aggregates observations of hand usage
across multiple temporal segments into a consistent
interpretation. \textsc{Cashew-RL} internalizes this reasoning, attending to key objects such as hands, chips, basket, and table, producing a concise, temporally grounded explanation that correctly tracks both hands. Figure~\ref{fig:process_examples} provides a detailed view of this iterative aggregation process for the first example.

\end{document}